\documentclass[10pt,journal,compsoc]{IEEEtran}

\ifCLASSOPTIONcompsoc
  \usepackage[nocompress]{cite}
\else
  \usepackage{cite}
  \usepackage{natbib}
\fi

\ifCLASSINFOpdf
\else
\fi

\usepackage[pdftex, colorlinks,linkcolor={red},anchorcolor={blue},citecolor={green},urlcolor={magenta}, backref=page]{hyperref}
\usepackage{amsmath}
\usepackage{amssymb}
\usepackage{bm}
\usepackage[dvipsnames]{xcolor}
\usepackage{graphicx}
\usepackage{subcaption}

\usepackage{booktabs}
\usepackage{multirow}
\usepackage{pifont} 

\begin{document}

\title{SODFormer: Streaming Object Detection with Transformer Using Events and Frames}

\author{Dianze~Li,
        Jianing~Li,~\IEEEmembership{Member,~IEEE}
        and~Yonghong~Tian,~\IEEEmembership{Fellow,~IEEE}
\IEEEcompsocitemizethanks{\IEEEcompsocthanksitem Dianze Li and Jianing Li are with the National Engineering Research Center for Visual Technology, School of Computer Science, Peking University, Beijing 100871, China.\protect ~E-mail: dianzeli@stu.pku.edu.cn, lijianing@pku.edu.cn.}

\IEEEcompsocitemizethanks{\IEEEcompsocthanksitem Yonghong Tian are with the National Engineering Research Center for Visual Technology, School of Computer Science, Peking University, Beijing 100871, China, and also with the Peng Cheng Laboratory, Shenzhen 518000, China.\protect ~E-mail: yhtian@pku.edu.cn.}

\thanks{Manuscript received September 23, 2022.}
\thanks{(Corresponding author: Yonghong Tian and Jianing Li).}}

\markboth{IEEE Transactions on Pattern Analysis and Machine Intelligence, 2022}%
{Li \MakeLowercase{\textit{et al.}}: SODFormer: Streaming Object Detection with Transformer Using Events and Frames}

\IEEEtitleabstractindextext{%
\begin{abstract}
DAVIS camera, streaming two complementary sensing modalities of asynchronous events and frames, has gradually been used to address major object detection challenges (e.g., fast motion blur and low-light). However, how to effectively leverage rich temporal cues and fuse two heterogeneous visual streams remains a challenging endeavor. To address this challenge, we propose a novel streaming object detector with Transformer, namely SODFormer, which first integrates events and frames to continuously detect objects in an asynchronous manner. Technically, we first build a large-scale multimodal neuromorphic object detection dataset (i.e., PKU-DAVIS-SOD) over 1080.1k manual labels. Then, we design a spatiotemporal Transformer architecture to detect objects via an end-to-end sequence prediction problem, where the novel temporal Transformer module leverages rich temporal cues from two visual streams to improve the detection performance. Finally, an asynchronous attention-based fusion module is proposed to integrate two heterogeneous sensing modalities and take complementary advantages from each end, which can be queried at any time to locate objects and break through the limited output frequency from synchronized frame-based fusion strategies. The results show that the proposed SODFormer outperforms four state-of-the-art methods and our eight baselines by a significant margin. We also show that our unifying framework works well even in cases where the conventional frame-based camera fails, e.g., high-speed motion and low-light conditions. Our dataset and code can be available at \url{https://github.com/dianzl/SODFormer}.
\end{abstract}

\begin{IEEEkeywords}
Neuromorphic Vision, Event Cameras, Object Detection, Transformer, Multimodal Fusion.
\end{IEEEkeywords}}

\maketitle

\IEEEdisplaynontitleabstractindextext


\IEEEpeerreviewmaketitle

\ifCLASSOPTIONcompsoc
\IEEEraisesectionheading{\section{Introduction}\label{sec:introduction}}
\else

\section{Introduction}
\label{sec:introduction}
\fi

\IEEEPARstart{O}{bject} detection~\cite{liu2020deep, long2020scene, oksuz2020imbalance}, one of the most fundamental and challenging topics, supports a wide range of computer vision tasks, such as autonomous driving, intelligent surveillance, robot vision, etc. In fact, with conventional frame-based cameras, object detection performance~\cite{sun2019fab, sayed2021improved, hu2020learning} has suffered from a significant drop in some challenging conditions (e.g., high-speed motion blur, low-light, and overexposure). A key question still remains: \emph{How to utilize a novel sensing paradigm that makes up for the limitations of conventional cameras?}

\begin{figure}[t]
	\begin{subfigure}[b]{0.49\linewidth}
		\centering
		\centerline{\includegraphics[height=2.05cm]{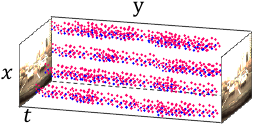}}
		\caption{Events and frames}
		\label{fig:1(a)}
	\end{subfigure}
	\begin{subfigure}[b]{0.49\linewidth}
		\centering
		\centerline{\includegraphics[height=1.85cm]{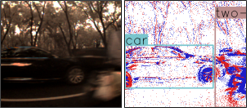}}
		\caption{Motion blur}
		\label{fig:1(b)}
	\end{subfigure}
	\vspace{0.15cm}
	
	\begin{subfigure}[b]{0.49\linewidth}
		\centering
		\centerline{\includegraphics[height=1.85cm]{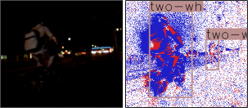}}
		\caption{Low-light condition}
		\label{fig:1(c)}
	\end{subfigure}
	\begin{subfigure}[b]{0.49\linewidth}
		\centering
		\centerline{\includegraphics[height=1.85cm]{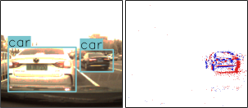}}
		\caption{Static scenes}
		\label{fig:1(d)}
	\end{subfigure}
	\caption{This work makes complementary use of events and frames to detect objects. (a) DAVIS~\cite{brandli2014240} outputs events and frames. (b)-(d) The left refers to frames, and the right denotes events. Note that, the event camera has the ability of high-speed and high dynamic range sensing, but it fails to capture static textures as a conventional camera does.}
	\label{fig:figure1}
	\vspace{-0.30cm}
\end{figure}

More recently, Dynamic and Active-Pixel Vision Sensor~\cite{brandli2014240, moeys2017sensitive} (i.e., DAVIS), namely a multimodal vision sensor, is designed in that spirit, combining a bio-inspired event camera and a conventional frame-based camera in the same pixel array. Its core is the novel event camera (i.e., DVS~\cite{lichtsteiner2008128}), which works differently from frame-based cameras and reacts to light changes by triggering asynchronous events. Since the event camera offers high temporal resolution ($\mu s$) and high dynamic range (HDR, up to 120 dB), it has brought a new perspective to address some limitations of conventional cameras in fast motion and challenging light scenarios. However, just as conventional frame-based cameras fail in extreme-light or high-speed motion blur scenarios, event cameras perform poorly in static or extremely slow-motion scenes. We refer to this phenomenon as unimodal degradation because it is mainly caused by the limitations of unimodal cameras (see Fig.~\ref{fig:figure1}). While event cameras are insensitive in static or extremely slow-motion scenes, frame-based cameras directly provide static fine textures (i.e., absolute brightness) conversely. Indeed, event cameras and frame-based cameras are complementary, which motivates the development of novel computer vision tasks (e.g., feature tracking~\cite{gehrig2020eklt}, depth estimation~\cite{gehrig2021combining}, and video reconstruction~\cite{tulyakov2021time}) by taking advantage of each end. Therefore, we aim at making complementary use of asynchronous events and frames to maximize object detection accuracy.

One problem is that most existing event-based object detectors~\cite{jiang2019mixed, li2019event, cao2021fusion, liu2021attention} run feed-forward models independently, leading to \emph{not utilizing rich temporal cues} in continuous visual streams. When isolated event image~\cite{maqueda2018event} may have issues involving small objects, occlusion, and out of focus, it is natural for the human visual system~\cite{dolan1997brain, huth2012continuous} to identify objects in the temporal dimension and then assign them together. In other words, the detection performance can be further improved via leveraging rich temporal cues from event streams and adjacent frames. Although the field of computer vision has witnessed significant achievements of CNNs and RNNs, they show poor ability in modeling long-term temporal dependencies. Consequently, the emerging Transformer is gaining more and more attention in object detection tasks~\cite{carion2020end, zhu2020deformable}. Thanks to the self-attention mechanism, Transformer~\cite{vaswani2017attention, khan2021transformers} is particularly suited to modeling long-term temporal dependencies for video sequence tasks~\cite{kim2018spatio, he2021end, zhang2021vidtr}. But, there's still no Transformer-based object detector to exploit temporal cues from events and frames. Meanwhile, the rare multimodal neuromorphic object detection datasets (e.g., PKU-DDD17-CAR~\cite{li2019event} and DAD~\cite{liu2021attention}) only provide isolated images and synchronized events, which means there is a lack of temporally long-term continuous and large-scale multimodal datasets including events, frames, and manual labels.

Another problem is that current fusion strategies (e.g., posting-processing~\cite{jiang2019mixed, li2019event}, concatenation~\cite{tomy2022fusing} and averaging~\cite{liu2021attention, li2020mdlatlrr}) for events and frames are \emph{inadequate in exploiting complementary information between two streams while only able to synchronously predict results with the frame rate, incurring the bottlenecks of performance improvements and inference frequency.} Since asynchronous events are sparse points in the spatiotemporal domain with higher temporal resolution compared to structured frames, the existing frame-based multimodal detectors cannot directly deal with two heterogeneous visual streams. As a result, some fusion strategies attempt to first split the continuous event streams into discrete image-like representations with the same sampling frequency of frames and then integrate two synchronized streams via the post-processing or feature aggregation operation. In short, one drawback is that these fusion strategies can't distinguish the degree of degradation of modalities and regions, therefore fail to eliminate the unimodal degradation thoroughly. The other drawback is that the inference frequency of these synchronized fusion strategies is limited by the sampling rate of conventional frames. Yet, there is no work to design an effective asynchronous multimodal fusion module for events and frames in the object detection task.

To address the aforementioned problems, this paper proposes a novel streaming object detector with Transformer, namely \textbf{\emph{SODFormer}}, which continuously detects objects in an asynchronous way by integrating events and frames. Actually, this work aims not to optimize Transformer-based object detectors (e.g., DETR~\cite{carion2020end}) on each isolated image. In contrast, our goal is to overcome \textbf{\emph{the following challenges}}: (i) \emph{\textbf{Lack of dataset} - How do we create a large-scale multimodal neuromorphic dataset for object detection}? (ii) \emph{\textbf{Temporal correlation} - How do we design a Transformer-based object detector that leverages rich temporal cues}? (iii) \emph{\textbf{Asynchronous fusion} - What is the unifying mechanism that takes advantages from two streams and achieves high inference frequency}? To this end, we first build a large-scale multimodal object detection dataset (i.e., PKU-DAVIS-SOD), which provides manual bounding boxes at a frequency of 25 Hz for 3 classes, yielding more than 1080.1k labels. Then, we propose a spatiotemporal Transformer that aggregates spatial information from continuous streams, which outputs the detection results via an end-to-end sequence prediction problem. Finally, an asynchronous attention-based fusion module is designed to effectively integrate two sensing modalities while eliminating unimodal degradation and breaking the limitation of frame rate. The results show that our SODFormer outperforms the state-of-the-art methods and our two single-modality baselines by a large margin. We also verify the efficacy of our SODFormer in fast motion and low-light scenarios.


To sum up, the main contributions of this work are summarized as follows:
\begin{itemize}
	\item We propose a novel Transformer-based framework for \emph{streaming object detection} (SODFormer), which first integrates events and frames via Transformer to continuously detect objects in an asynchronous manner.
	\item We design an effective \emph{temporal Transformer module}, which leverages rich temporal cues from two visual streams to improve the object detection performance.
	\item We develop an \emph{asynchronous attention-based fusion module} that takes complementary advantages from each end, eliminates unimodal degradation and overcomes the limited inference frequency from frame rate.
	\item We establish a \emph{large-scale and temporally long-term multimodal neuromorphic dataset} (i.e., PKU-DAVIS-SOD), which will open an opportunity for the research of this challenging problem.
\end{itemize}

The rest of the paper is organized as follows. Section~\ref{related} reviews prior works. We describe how to build a competitive dataset in Section~\ref{dataset}. Section~\ref{problem} introduces the camera working principle and formulates the novel problem. Section~\ref{method} presents our solution of streaming object detection with Transformer. Finally, Section~\ref{experiment} analyzes the performance of the proposed method, and some discussions are present in Section~\ref{discussion}, while some conclusions are drawn in Section~\ref{conclusion}.

\section{Related Work}
\label{related}
This section first reviews neuromorphic object detection datasets (Section~\ref{review-datasets}) and the corresponding object detectors (Section~\ref{review-approaches}), followed by an overview of object detectors with Transformer (Section~\ref{transformers}) and a survey of fusion approaches for events and frames (Section~\ref{review_fusion}).

\subsection{Neuromorphic Object Detection Datasets} \label{review-datasets}
The publicly available object detection datasets utilizing event cameras have a limited amount~\cite{gallego2020event, liu2019event, roy2019towards, chen2020event}. Gen1 Detection dataset~\cite{de2020large} and 1Mpx Detection dataset~\cite{perot2020learning} provide large-scale annotations for object detection using event cameras. However, it may be difficult to obtain fine textures and achieve high-precision object detection by only using DVS events. Although event-based simulators (e.g., ESIM~\cite{rebecq2018esim}, V2E~\cite{hu2021v2e}, and RS~\cite{kang2021retinomorphic}) can directly convert video object datasets to the neuromorphic domain, those transformed strategies fail to reflect realistic high-speed motion or extreme light scenarios, which are exactly what event cameras are good at. Besides, PKU-DDD17-CAR dataset~\cite{li2019event} and DAD dataset~\cite{liu2021attention} provide 3155 and 6247 isolated hybrid sequences, respectively. Nevertheless, these two small-scale datasets are not continuous streams modeling temporal dependencies. Therefore, this work aims to establish a large-scale and temporally long-term multimodal object detection dataset involving challenging scenarios.

\subsection{Neuromorphic Object Detection Approaches} \label{review-approaches}
The existing object detectors using event cameras can be roughly divided into two categories. The first category~\cite{iacono2018towards, chen2018pseudo, cannici2019event, chen2019multi, ryan2021real, li2022asynchronous, xiang2022temporal, wang2023dual, gehrig2023recurrent} is the single-modality that only processes asynchronous events. These methods usually first convert asynchronous events into 2D image-like representations~\cite{gehrig2019end}, and then input them into frame-based detectors (e.g., YOLOs~\cite{redmon2018yolov3}). Although only utilizing dynamic events can achieve satisfactory performance in some specific scenes, it becomes clear that high-precision detection requires static fine textures.

The second category~\cite{liu2016combined, li2019event, jiang2019mixed, hu2020learning, cao2021fusion, liu2021attention, tomy2022fusing, el2022high, messikommer2022bridging} refers to the multi-modality that combines multiple visual streams. For example, some works~\cite{liu2016combined, li2019event, jiang2019mixed} first detect objects on each isolated frame or event temporal bin, and then merge the detection results of two modalities by post-processing (e.g., NMS~\cite{hosang2017learning}). Besides, a grafting algorithm~\cite{hu2020learning} is proposed to integrate events and thermal images. Some attention-based aggregation operations~\cite{cao2021fusion, tomy2022fusing} are designed to fuse each isolated frame and event temporal bin from the PKU-DDD17-CAR~\cite{li2019event} dataset, while others~\cite{liu2021attention,li2020mdlatlrr} average the features from each modality. However, these joint frameworks have not explored rich temporal cues from continuous visual streams. Moreover, these post-processing or aggregation fusion operations are hard to capture global context across two sensing modalities, while averaging strategy cannot eliminate unimodal degradation thoroughly. Thus, we design a streaming object detector that aims at leveraging rich temporal cues and making fully complementary use of two visual streams.

\subsection{Object Detection with Transformer} \label{transformers}
Transformer, an attention-based architecture, is first introduced by~\cite{vaswani2017attention} for machine translation. Its core attention mechanism~\cite{neu2015dzm} scans through each element of a sequence and updates it by aggregating information from the whole sequence with different attention weights. The global computation makes Transformers perform better than RNNs on long sequences. Until now, Transformers have been migrated to computer vision and replaced RNNs in many problems. In video object detection, while RNNs~\cite{ric2013ros, fas2015ros, ren2017fas} have achieved great success, they require meticulously hand-designed components (e.g., anchor generation). To simplify these pipelines, DETR~\cite{carion2020end}, an end-to-end Transformer-based object detector, is proposed. It is worth mentioning that DETR~\cite{carion2020end} is the first Transformer-based object detector via a sequence predicting model. While DETR largely simplifies the classical CNN-based detection paradigm, it suffers from very slow convergence and relatively low performance at detecting small objects. More recently, a few approaches attempt to design optimized architectures to help detect small objects and speed up training. For instance, Deformable DETR~\cite{zhu2020deformable} adopts the multi-scale deformable attention module that achieves better performance than DETR with 10$\times$ fewer training epochs. PnP-DETR~\cite{wang2021pnp} significantly reduces spatial redundancy and achieves more efficient computation via a two-step poll-and-pool sampling module. However, most works operate on each isolated image and do not exploit temporal cues from continuous visual streams. Inspired by the ability of Transformer to model long-term dependencies in video sequence tasks~\cite{kim2018spatio, he2021end, zhang2021vidtr}, we propose a temporal Transformer model to leverage rich temporal cues from continuous events and adjacent frames.

\subsection{Multimodal Fusion for Events and Frames} \label{review_fusion}
Some computer vision tasks (e.g., video reconstruction~\cite{tulyakov2021time, shang2021bringing, zhu2021neuspike, duan2021guided}, object detection~\cite{li2019event, jiang2019mixed}, object tracking~\cite{zhang2021object, wang2021visevent}, depth estimation~\cite{gehrig2021combining}, and SLAM~\cite{vidal2018ultimate, zuo2022devo, gao2022vector}) have sought to integrate two complementary modalities of events and frames. For example, JDF~\cite{li2019event} adopts the Dempster-Shafer theory to fuse events and frames for object detection. A recurrent asynchronous multimodal network~\cite{gehrig2021combining} is introduced to fuse events and frames for depth estimation. In fact, most fusion strategies (e.g., post-processing, concatenation, and averaging) can't distinguish the degradation degree of different modalities and regions, resulting in the failure to eliminate unimodal degradation completely (see Section~\ref{review-approaches}). More importantly, most fusion strategies are synchronized with event and frame streams, which causes the joint output frequency to be limited by the sampling rate of conventional frames. Obviously, the limited inference frequency can't meet the needs of fast object detection in real-time high-speed scenarios, so we seek to utilize the high temporal resolution advantage of event cameras and make use of events and frames. In RAMNet~\cite{gehrig2021combining}, RNN is utilized to introduce an asynchronous fusion method that allows decoding the task variable at any time. Nevertheless, to the best of our knowledge, there are no prior attention-based works involving asynchronous fusion strategy for events and frames in the object detection task. Thus, we design a novel asynchronous attention-based fusion module that eliminates unimodal degradation in two modalities with high inference frequency.

\begin{figure}
	\begin{subfigure}[b]{0.5\linewidth}
		\centering
		\centerline{\includegraphics[height=1.80cm]{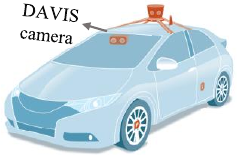}}
		\caption{Recording platform}
		\label{fig:2(a)}
	\end{subfigure}
	\begin{subfigure}[b]{0.5\linewidth}
		\centering
		\centerline{\includegraphics[height=1.80cm]{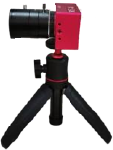}}
		\caption{DAVIS346 camera}
		\label{fig:2(b)}
	\end{subfigure}
	\vspace{-0.30cm}
	\caption{Experimental setup. A DAVIS346 camera is installed on the front windshield of the driving car.}
	\label{fig:figure2}
	\vspace{-0.30cm}
\end{figure}

\section{PKU-DAVIS-SOD Dataset}
\label{dataset}

This section first presents the details of how to build our dataset (Section~\ref{collection}). Then, we give detailed statistics to better understand the newly built dataset (Section~\ref{static}). Finally, we make a comparison of related datasets (Section~\ref{compare}).

\subsection{Data Collection and Annotation} \label{collection}
The goal of this dataset is to offer a dedicated platform for the training and evaluation of streaming object detection algorithms. Thus, a novel multimodal vision sensor (i.e., DAVIS346, resolution 346$\times$260) is used to record multiple hybrid sequences in a driving car (see Fig.~\ref{fig:figure2}). The event camera simultaneously outputs events with high temporal resolution and conventional frames with 25 FPS. Mostly, we fix the DAVIS346 in a driving car (see Fig.~\ref{fig:figure2}(\subref{fig:2(a)})), and record the sequences while the car is driving on city roads. Nevertheless, it is difficult to capture high-speed motion blur owing to the relative speed between vehicles on city roads. For the convenience of acquiring high-speed objects, we additionally provide some sequences in which the camera is set at the flank of the road. The raw recordings consider velocity distribution, light condition, category diversity, object scale, etc. To provide manual bounding boxes in challenging scenarios (e.g., high-speed motion blur and low-light), grayscale images are reconstructed from asynchronous events using E2VID~\cite{rebecq2019events} at 25 FPS when RGB frames are of low quality. When the objects are not visible in all three modalities (i.e., RGB frames, event images, and reconstructed frames), they can still be labeled because our PKU-DAVIS-SOD dataset provides continuous visual streams, and we can obtain information about the unclear objects from nearby images. Furthermore, some unclear objects in a single image may be visible in a piece of video. After the temporal calibration, we first select three common and important object classes (i.e., car, pedestrian, and two-wheeler) in our daily life. Then, all bounding boxes are annotated from RGB frames or synchronized reconstructed images by a well-trained professional team.

\begin{table}
	\begin{center}
	\caption{The number of labeled frames and bounding boxes in each set of our PKU-DAVIS-SOD dataset.}
	\label{tab:table1}
	\setlength{\tabcolsep}{1.80mm}{
		\begin{tabular}{l cc cc cc cc}
			\toprule
			\multirow{2}*{Name} & \multicolumn{2}{c}{Normal} & \multicolumn{2}{c}{Motion blur} & \multicolumn{2}{c}{Low-light}\\ 
			\cline{2-3} \cline{4-5} \cline{6-7}& frames & boxes  & frames & boxes & frames & boxes \\ 
			\hline
			Training & 115104 & 528944 & 31587 & 87056 & 22826 & 55343 \\
			Validation & 31786 & 155778 & 9763 & 24828 & 8892 & 14143 \\
			Test & 35459 & 150492 & 11859 & 44757 & 8729 & 18807 \\
			\hline
			Total & 182349 & 835214 & 53209 & 156641 & 40447 & 88293 \\
			\bottomrule
	\end{tabular}}
	\end{center}
	\vspace{-0.20cm}
\end{table}

\subsection{Dataset Statistics} \label{static}
PKU-DAVIS-SOD dataset provides 220 hybrid driving sequences and labels at a frequency of 25 Hz. As a result, this dataset contains 276k timestamps (i.e., labeled frames) and 1080.1k bounding boxes in total. Afterward, we split them into 671.3k for training, 194.7k for validation, and 214.1k for testing. Table~\ref{tab:table1} shows the number of bounding boxes in each set. Besides, we further analyze the attributes of the newly built dataset from the four following perspectives.

\begin{figure}[t]
	\begin{subfigure}[b]{0.49\linewidth}
		\centering
		\centerline{\includegraphics[height=2.50cm]{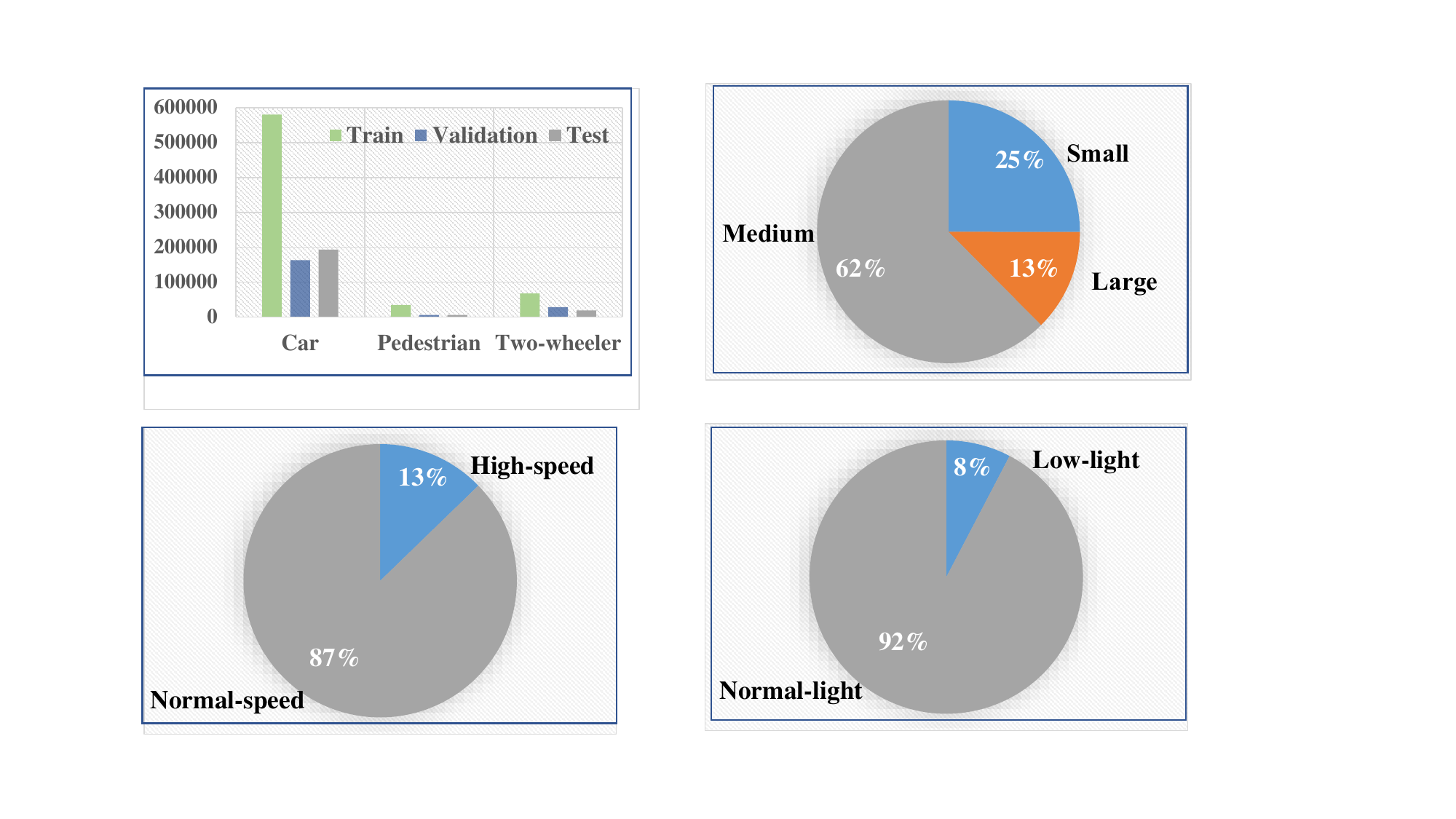}}
		\caption{Category diversity}
		\label{fig:3(a)}
	\end{subfigure}
	\begin{subfigure}[b]{0.49\linewidth}
		\centering
		\centerline{\includegraphics[height=2.48cm]{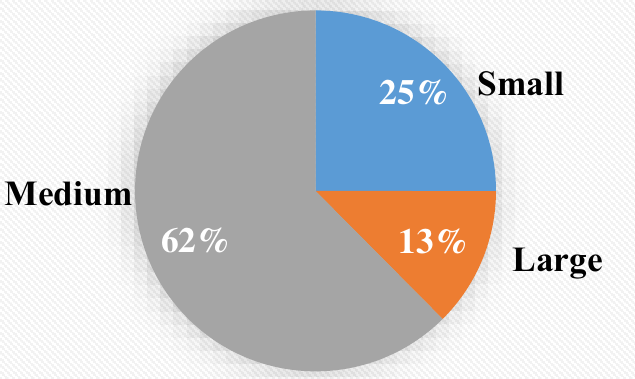}}
		\caption{Object scale}
		\label{fig:3(b)}
	\end{subfigure}
	\vspace{0.20cm}
	
	\begin{subfigure}[b]{0.49\linewidth}
		\centering
		\centerline{\includegraphics[height=2.48cm]{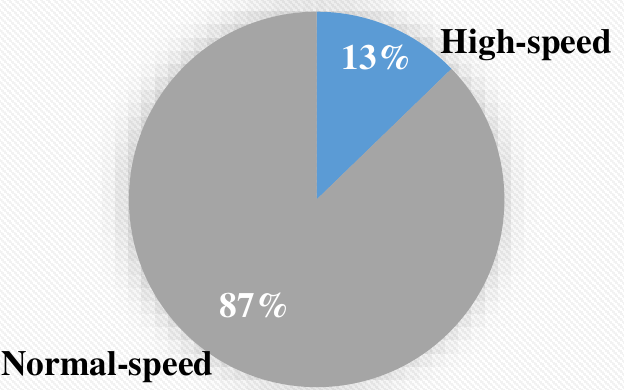}}
		\caption{Velocity distribution}
		\label{fig:3(c)}
	\end{subfigure}
	\begin{subfigure}[b]{0.49\linewidth}
		\centering
		\centerline{\includegraphics[height=2.48cm]{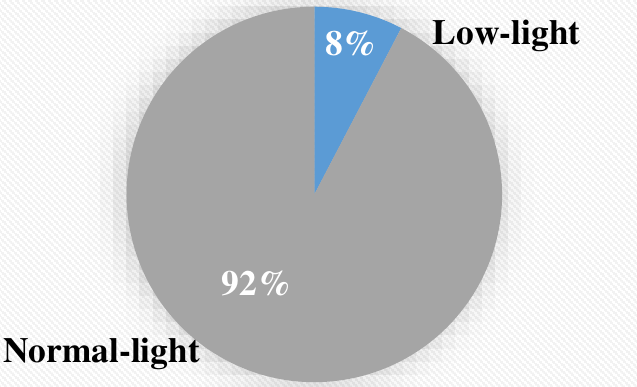}}
		\caption{Light intensity}
		\label{fig:3(d)}
	\end{subfigure}
	\caption{Data statics in our PKU-DAVIS-SOD dataset. (a) The distributions of three types of objects (i.e., car, pedestrian, and two-wheeler). (b)-(d) The proportions of object scales, moving speeds, and light intensities.}
	\label{fig:figure3}
	\vspace{-0.30cm}
\end{figure}

\emph{Category Diversity}. Fig.~\ref{fig:figure3}(\subref{fig:3(a)}) displays the number distributions of three types of labeled objects (i.e., car, pedestrian, and two-wheeler) in our PKU-DAVIS-SOD dataset. We can find that the numbers of cars, pedestrians, and two-wheelers in each set (i.e., training, validation, and testing) are (580340, 162894, 193118), (34744, 5800, 5680), and (67599, 28400, 19144), respectively. For intuition, the ratios of the above numbers are approximately 3.5 : 1 : 1.2, 6 : 1 : 1, and 3.5 : 1.5 : 1, respectively.

\emph{Object Scale}. Fig.~\ref{fig:figure3}(\subref{fig:3(b)}) shows the proportions or object scales in our PKU-DAVIS-SOD dataset. The object height is used to reflect its scale since the height and the scale are closely related in driving scenes~\cite{pang2020tju}. We broadly classify all objects into three scales (i.e., large, medium, and small). The medium scale varies from 20 pixels to 80 pixels, the small refers to the object less than 20 pixels in height, and the large scale means the object height is greater than 80 pixels. 

\emph{Velocity Distribution}. As illustrated in Fig.~\ref{fig:figure3}(\subref{fig:3(c)}), we divide the motion speed level into normal-speed and high-speed. Event camera, offering high temporal resolution, can capture high-speed moving objects, such as a rushing car. On the contrary, RGB frames may result in motion blur due to the limited frame rate. In the process of dividing the dataset, we use video as the basic unit. If a sequence contains lots of motion blur scenarios, it is classified as high-speed, otherwise, it would be classified as normal-speed instead. Fig.~\ref{fig:figure3}(\subref{fig:3(c)}) shows that the proportion of high-speed scenarios is 13\%, and the remaining involving normal-speed objects is 87\%.

\emph{Light Intensity}. Since the raw visual streams are recorded from daytime to night, it covers the diversity in illumination variance. The sequences belonging to the low-light scenario are generally collected in low-light conditions (night, rainy days, tunnels, etc.) and their RGB frames consist mainly of low-light scenes. Thus, we can easily judge the light condition by viewing RGB frames with our eyes. Fig.~\ref{fig:figure3}(\subref{fig:3(d)}) illustrates that the proportions of normal light and low-light are 92\% and 8\%, respectively.

\begin{figure*}[t]
	\begin{subfigure}[b]{0.247\linewidth}
		\centering
		\centerline{\includegraphics[height=4.85cm]{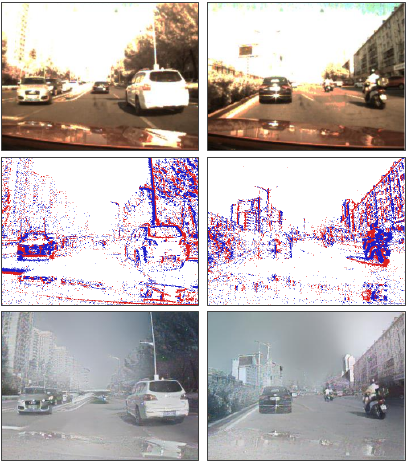}}
		\caption{Category diversity}
		\label{fig:4(a)}
	\end{subfigure}
	\begin{subfigure}[b]{0.247\linewidth}
		\centering
		\centerline{\includegraphics[height=4.85cm]{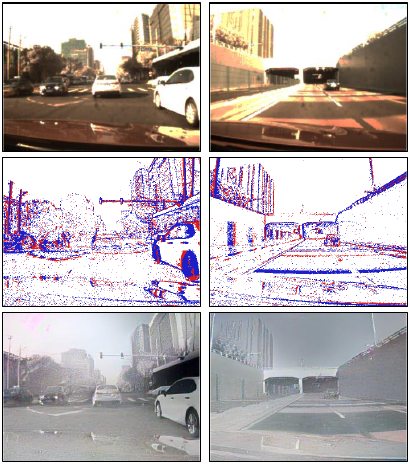}}
		\caption{Object scale}
		\label{fig:4(b)}
	\end{subfigure}
	\begin{subfigure}[b]{0.247\linewidth}
		\centering
		\centerline{\includegraphics[height=4.85cm]{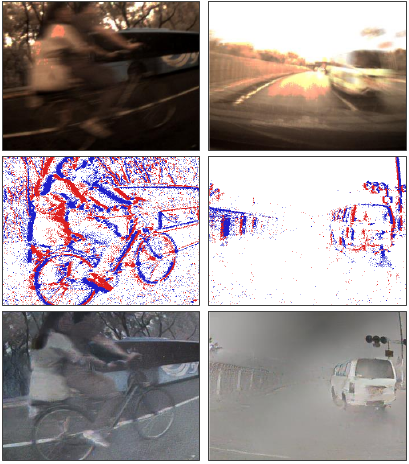}}
		\caption{Velocity distribution}
		\label{fig:4(c)}
	\end{subfigure}
	\begin{subfigure}[b]{0.247\linewidth}
		\centering
		\centerline{\includegraphics[height=4.85cm]{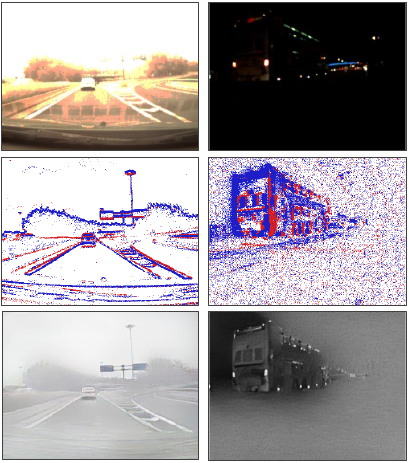}}
		\caption{Light change}
		\label{fig:4(d)}
	\end{subfigure}
	\caption{Representative samples of our PKU-DAVIS-SOD dataset. The three rows from top to bottom refer to RGB frame, event images, and DVS reconstructed images from asynchronous events using E2VID~\cite{rebecq2019events}.}
	\label{fig:figure4}
\end{figure*}

For better visualization, we show representative samples (see Fig.~\ref{fig:figure4}) including RGB frames, event images, and DVS reconstructed images, which cover the category diversity, object scale, velocity, and light change.

\begin{table*}
	\caption{Comparison with related object detection datasets using event cameras. Note that, our PKU-DAVIS-SOD dataset is the first large-scale multimodal neuromorphic object detection dataset that provides temporally long-term asynchronous events, conventional frames, and manual labels at 25 Hz.}
	\label{tab:table2}
	\vspace{-0.35cm}
	\begin{center}
		\setlength{\tabcolsep}{1.60mm}{
			\begin{tabular}{lcccccccccc}
				\toprule
				Dataset & Year  & Venue    & Resolution & Modality  & Classes  & Boxes & Label & Frequency  & High-speed & Low-light  \\
				\hline
				PKU-DDD17-CAR~\cite{li2019event} & 2019 & ICME   & 346$\times$260 & Events, Frames   & 1  & 3155 & Manual  & 1 Hz  & \ding{55} & \ding{51}  \\
				TUM-Pedestrian~\cite{jiang2019mixed} & 2019  & ICRA & 240$\times$180  & Events, Frames  & 1  & 9203 & Manual  & 1 Hz  & \ding{55} & \ding{55}   \\
				Pedestrian Detection~\cite{chen2019multi} & 2019 & FNR & 240$\times$180 & Events  &1  & 28109 & Manual & 1 Hz & \ding{55} & \ding{55}   \\
				Gen1 Detection~\cite{de2020large} & 2020  & arXiv   & 304$\times$240 & Events  & 2  & 255k & Pseudo & 1, 4 Hz  & \ding{55} & \ding{51} \\
				1 Mpx Detection~\cite{perot2020learning} &  2020 & NIPS  & 1280$\times$720 & Events   & 3  & 25M & Pseudo & 60 Hz  & \ding{55} & \ding{51}  \\
				DAD~\cite{liu2021attention} & 2021  & ICIP & 346$\times$260 & Events, Frames & 1  & 6427 & Manual & 1 Hz  & \ding{55} & \ding{51}  \\
				\bottomrule
				\textbf{PKU-DAVIS-SOD} & 2022  & Ours  & 346$\times$260 & \textbf{Events, Frames}  & \textbf{3}  & \textbf{1080.1k} & \textbf{Manual} & \textbf{25 Hz}  & \ding{51} & \ding{51}  \\
				\bottomrule
		\end{tabular}}
	\end{center}
	\vspace{-0.30cm}
\end{table*}

\subsection{Comparison with Other Datasets} \label{compare}
To clarify the advancements of the newly built dataset, we compare it with some related object detection datasets using event cameras in Table~\ref{tab:table2}. Apparently, our PKU-DAVIS-SOD dataset is the first large-scale and open-source multimodal neuromorphic object detection dataset~\footnote[1]{\url{https://www.pkuml.org/research/pku-davis-sod-dataset.html}}. In contrast, two publicly large-scale datasets (i.e., Gen1 Detection~\cite{de2020large} and 1 Mpx Detection~\cite{perot2020learning}) only provide temporally long-term event streams, thus it is difficult for them to serve high-precision object detection, especially in static or extremely slow motion scenarios. What's more, TUM-Pedestrian dataset~\cite{jiang2019mixed} and Pedestrian Detection dataset~\cite{chen2019multi} provide event streams for pedestrian detection, but they have not yet been publicly available. DAD dataset~\cite{liu2021attention}, followed by PKU-DDD17-CAR dataset~\cite{li2019event}, offers isolated frames and event temporal bins rather than temporally continuous visual streams.

All in all, such a novel bio-inspired multimodal camera and professional design with high labor intensity enable our PKU-DAVIS-SOD to be a competitive dataset with \emph{\textbf{multiple characteristics}}: (i) \emph{High temporal sampling resolution with 12 Meps from event streams}; (ii) \emph{High dynamic range property with 120 dB from event streams}, (iii) \emph{Two temporally long-term visual streams with labels at 25 Hz}, (iv) \emph{Real-world scenarios with abundant diversities in object category, object scale, moving velocity, and light change}.

\section{Preliminary and Problem Definition}
\label{problem}

This section first presents the details of DAVIS camera working principle (Section~\ref{principle}). Then, we formulate the definition of streaming object detection (Section~\ref{streaming}).

\subsection{DAVIS Sampling Principle} \label{principle}
DAVIS is a multimodal vision sensor that simultaneously outputs two complementary modalities of asynchronous events and frames under the same pixel array. Due to the complementarity of events and frames, the output of the DAVIS camera contains more information, making the DAVIS camera naturally surpass unimodal sensors like conventional RGB cameras and event-based cameras. In particular, the core event camera~\cite{posch2014retinomorphic}, namely DVS, has independent pixels that react to changes in light intensity $R(\bm{u},t)$ with a stream of events. More specifically, an event $e_{n}$ is a four-attribute tuple $(x_{n}, y_{n}, t_{n}, p_{n})$ using addressing event representation (AER)~\cite{lazzaro1993silicon}, triggered for the pixel $\bm{u}=(x_{n}, y_{n})$ at the timestamp $t_{n}$ when the log-intensity changes over the pre-defined threshold $\theta_{th}$. This process can be depicted as:
\begin{eqnarray}
\text{ln} R(\bm{u}_{n}, t_{n})-\text{ln} R(\bm{u}_{n}, t_{n}-\Delta t_{n}) =p_{n} \theta_{th},
\end{eqnarray}
where $\Delta t_{n}$ is the temporal sampling interval at a pixel, the polarity $p_{n} \in \left\lbrace -1,1\right\rbrace $ refers to whether the brightness is decreasing or increasing.

Intuitively, asynchronous events appear as sparse and discrete points~\cite{li2021asynchronous} in the spatiotemporal domain, which can be described as follows:
\begin{eqnarray}
S\left(x,y,t\right)=\sum_{n=1}^{N_{e}}  p_{n}\delta\left(x-x_{n},y-y_{n},t-t_{n}\right),
\end{eqnarray}
where $N_{e}$ is the event number during the spatiotemporal window. $\delta\left(\cdot\right)$ refers to the Dirac delta function, with  $\int$$\delta\left(t\right)dt=1$ and $\delta\left(t\right)=0, \forall$$t\neq0$.

\subsection{Streaming Object Detection} \label{streaming}
Let  $S(x,y,t)$ and $I$=$\left\lbrace I_{1}, ..., I_{N}\right\rbrace $ are two complementary modalities of asynchronous events and frame sequence from a DAVIS camera. In general, to make the asynchronous events compatible with deep learning methods, a continuous event stream needs to be divided into $K$ event temporal bins $S=\left\lbrace S_{1}, ..., S_{K}\right\rbrace $. For the current timestamp $t_{i}$, the spatiotemporal location information of objects (i.e., bounding boxes) can be calculated using adjacent frames $\left\lbrace  I_{i-n}, ..., I_{i} \right\rbrace, n \in [0, N]$ and multiple temporal bins $ \left\lbrace S_{i-k}, ..., S_{i} \right\rbrace, k \in [0, K]$, it can be formulated by:
\begin{eqnarray}
B_{i}= \mathcal{D} \left( \left\lbrace  I_{i-n}, ..., I_{i} \right\rbrace,  \left\lbrace S_{i-k}, ..., S_{i} \right\rbrace  \right) ,
\end{eqnarray}
where $B_{i}=\left\lbrace  \left( x_{i,j}, y_{i,j}, w_{i,j}, h_{i,j}, c_{i,j}, p_{i,j}, t_{i} \right) \right\rbrace_{j\in[1,J]}$ is a list of $J$ bounding boxes in the timestamp $t_{i}$. More specifically, $(w_{i,j}, h_{i,j})$ are the width and the height of the $j$ bounding box, $(x_{i,j}, y_{i,j})$ are the corresponding upper-left coordinates, $c_{i,j}$ and $p_{i,j}$ are the predicted class and the confidence score of the $j$ bounding box, respectively. The function $\mathcal{D}$ is the proposed \textbf{\emph{streaming object detector}}, which can leverage rich temporal cues from $n+1$ adjacent frames and $k+1$ temporal bins. The parameters $n$ and $k$ determine the length of mining temporal information and affect the fusion strategy between two heterogeneous visual streams.  

Given the ground-truth $\bar{B}=\left\lbrace \bar{B}_{1}, ..., \bar{B}_{N}\right\rbrace $, we aim at making the output $B=\left\lbrace B_{1}, ..., B_{N}\right\rbrace $ of the optimized detector $\hat{\mathcal{D}}$ to fit $\bar{B}$ as much as possible, and it can formulate the following minimization problem as:
\begin{eqnarray}
\hat{\mathcal{D}}=\mathop{\arg \min}_{\mathcal{D}}\frac{1}{N} \sum\limits_{n=1}^{N}\mathcal{L_{\mathcal{D}}}(B_{i}, \bar{B}_{i}),
\end{eqnarray}
where $\mathcal{L_{\mathcal{D}}}$ is the designed loss function of the proposed streaming object detector.

Note that, this novel streaming object detector using events and frames works differently from the feed-forward object detection framework (e.g., YOLOs~\cite{redmon2018yolov3} and DETR~\cite{carion2020end}). It has two unique properties, which exactly account for the nomenclature of "streaming": (i) Objects can be accurately detected via leveraging rich temporal cues; (ii) Event streams offering high temporal resolution can meet the need of detecting objects at any time among continuous visual steams and overcome the limited inference frequency from conventional frames. Consequently, following similar works using point clouds~\cite{wei2020stream, chen2021polar} and video streams~\cite{Yang2022StreamYOLORO}, we call our proposed detector as streaming object detector.

\begin{figure*}[htbp]
	\centering
	\centerline{\includegraphics[width=\linewidth]{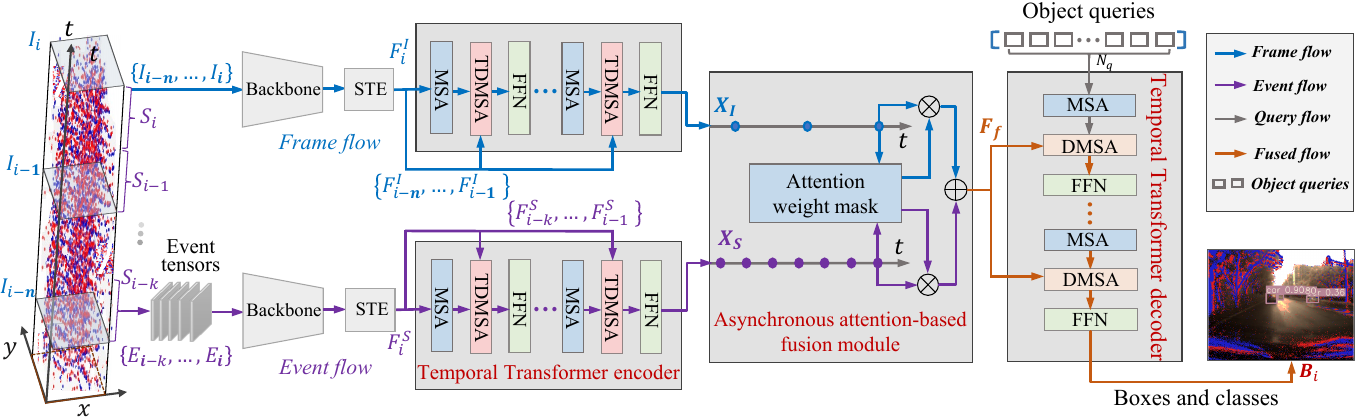}}
	\vspace{-0.15cm}
	\caption{The pipeline of the proposed \emph{streaming object detection with Transformer (SODFormer)}. The event stream is first split into event temporal bins and then encoded into event embeddings~\cite{gehrig2019end}. Then, we use a modal-specific shared backbone (e.g., ResNet50~\cite{he2016deep}) to extract features from frames and event embeddings, respectively. Meanwhile, our temporal Transformer encoder links the outputs of the spatial Transformer encoder (STE) for each stream. Our asynchronous attention-based fusion module exploits the attention mechanism to generate a fused representation. Finally, the designed temporal Transformer decoder integrates object queries and the fused flow to predict the final bounding boxes.}
	\vspace{-0.10cm}
	\label{fig:figure5}
\end{figure*}

\section{Methodology}
\label{method}

In this section, we first give an overview of our approach (Section~\ref{overview}). Besides, we briefly revisit the event representation (Section~\ref{representation}) and the spatial Transformer (Section~\ref{spatial}). Then, we present the details of how to exploit temporal cues from continuous visual streams (Section~\ref{temporal}). Finally, we design an asynchronous attention-based fusion strategy for events and frames (Section~\ref{fusion}).

\subsection{Network Overview}
\label{overview}
This work aims at designing a novel streaming object detector with Transformer, termed \emph{\textbf{SODFormer}}, which continuously detects objects in an asynchronous way by integrating events and frames. As illustrated in Fig.~\ref{fig:figure5}, our framework consists of four modules: \emph{event representation}, \emph{spatial Transformer}, \emph{\textbf{temporal Transformer}}, and \emph{\textbf{asynchronous attention-based fusion module}}. More precisely, to make asynchronous events compatible with deep learning methods, the continuous event stream is first divided into event temporal bins $S=\left\lbrace S_{1}, ..., S_{K}\right\rbrace $, and each bin $S_{i}$ can be converted into a 2D image-like representation $E_{i}$ (i.e., event embeddings~\cite{gehrig2019end}). Then, the spatial Transformer adopts the main structures of the Deformable DETR~\cite{zhu2020deformable} to extract feature representations from each sensing modality, which involves a feature extraction backbone (e.g., ResNet50~\cite{he2016deep}) and the spatial Transformer encoder (STE). Besides, the proposed temporal Transformer contains the temporal Deformable Transformer encoder (TDTE) and temporal Deformable Transformer decoder (TDTD). The TDTE assigns the outputs of STE in the temporal dimension, which can improve the accuracy of object detection by leveraging rich temporal information. Meanwhile, the proposed asynchronous attention-based fusion module exploits the attention mechanism to generate a fused feature, which can eliminate unimodal degradation in two modalities with a high inference frequency. Finally, The TDTD integrates object queries and the fused feature to predict the bounding boxes $B_{i}$.

\subsection{Event Representation}
\label{representation}
When using deep learning methods to process asynchronous events, spatiotemporal point sets are usually converted into successive measurements. In general, the kernel function~\cite{gehrig2019end} is usually used to map the event temporal bin $S_{i}$ into an event embedding $E_{i}$, which should ideally exploit the spatiotemporal information from asynchronous events. As a result, we can formulate this mapping as follows:
\begin{equation}
    \bm{E}_i = \sum_{e_n\in \Delta T} \mathcal{K}(x - x_n, y - y_n, t - t_n),
\end{equation}
where the designed kernel function $\mathcal{K} (x, y, t)$ can be deep neural networks or handcrafted operations.

In this work, we attempt to encode the event temporal bin into three existing event representations (i.e., event images~\cite{maqueda2018event}, voxel grids~\cite{zhu2019unsupervised} and sigmoid representation~\cite{chen2018pseudo}) owning to an accuracy-speed trade-off. Actually, any event representation can be an alternative because our SODFormer provides a generic interface accepting various input types of the event-based object detector.

\subsection{Revisiting Spatial Transformer}
\label{spatial}

In DETR~\cite{carion2020end}, the multi-head self-attention (MSA) combines multiple attention heads in parallel to increase the feature diversity. To be concrete, the feature maps are first extracted via CNN-based backbone. Then, $X_q$ and $X_k$ are derived from these feature maps as query and key-value in attention mechanism with both sizes of $(HW, d)$. Let $q$ refer to a query element with the feature map $X_{q}$, and $k$ indexes a key-value element with the feature map $X_{k}$. The MSA operation integrates the outputs of $M$ attention heads as:
\begin{equation}
\text {MSA}\left(X_{q}, X_{k}\right)=\sum_{m=1}^{M} \boldsymbol{W}_{m}\left[\sum A_{m q k} \cdot \boldsymbol{W}_{m}^{\prime} {X}_{k}\right],
\end{equation}
where $m$ indexes the single attention head, $\boldsymbol{W}_{m}$ and $\boldsymbol{W}_{m}^{\prime}$ are learnable weights. $A_{m q k}=\exp ( \frac{ \boldsymbol{X}_{q}^{T} \boldsymbol{W}_{mq}^{T} \boldsymbol{W}_{mk} \boldsymbol{X}_{k}}{\sqrt{d_{k}}})$ is the attention weight, in which $\boldsymbol{W}_{mq}$ and $\boldsymbol{W}_{mk}$ are also learnable weights, $d_{k}$ is a scaling factor.

To achieve fast convergence and high computation efficiency, Deformable DETR~\cite{zhu2020deformable} designs a deformable multi-head self-attention (DMSA) operation to attend the local $L$ sampling points instead of all pixels in feature map $X$. Specifically, DMSA first determines the corresponding locations of each query in other feature maps, which we refer to as reference points, and then adds the learnable sampling offsets to reference points to obtain the locations of sampling points. It can be described as:
\begin{equation}\label{dmsa}
\begin{aligned}
\text{DMSA}\left(X_{q}, X, P_{q}\right)=& \sum_{m=1}^{M} \boldsymbol{W}_{m}\left[\sum_{l=1}^{L} A_{m l q} \cdot \right.\\
&\left. \boldsymbol{W}_{m}^{\prime} X (P_{q}+\Delta P_{mlq})\right],
\end{aligned}
\end{equation}
where $P_{q}$ is a 2D reference point, $\Delta P_{mlq}$ denotes the sampling offset relative to $P_{q}$, and $A_{m l q}$ is the learnable attention weight of the $l^{th}$ point in $m^{th}$ self-attention head.

In this study, we adopt the encoder of the Deformable DETR as our spatial Transformer encoder (STE) to extract features from each visual stream. Meanwhile, we further extend the spatial DMSA strategy to the temporal domain (Section~\ref{temporal}), which leverages rich temporal cues from continuous event stream and adjacent frames.

\subsection{Temporal Transformer}
\label{temporal}

The temporal Transformer aggregates multiple spatial feature maps from the spatial Transformer and generates the predicted bounding boxes. It includes two main modules: temporal deformable Transformer encoder (TDTE) (Section~\ref{temporal_encoder}) and temporal deformable Transformer decoder (TDTD) (Section~\ref{temporal_decoder}).

\subsubsection{Temporal Deformable Transformer Encoder}
\label{temporal_encoder}

Our TDTE aims at aggregating multiple spatial feature maps from the spatial Transformer encoder (STE) using the temporal deformable multi-head self-attention (TDMSA) operation and then generating a spatiotemporal representation via the designed multiple stacked blocks.

\begin{figure}[t]
	\centerline{\includegraphics[width=\linewidth]{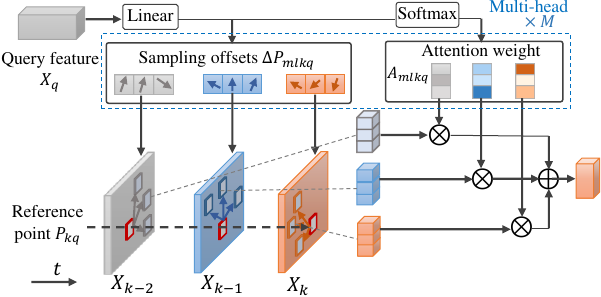}}
	\caption{The architecture of the proposed temporal deformable multi-head self-attention (TDMSA).} 
	\label{fig:figure6}
	\vspace{-0.0cm}
\end{figure}

The core idea of the proposed TDMSA operation is that the temporal deformable attention only attends to the local sampling points in the spatial domain and aggregates all sampling points in the temporal domain. Obviously, TDMSA directly extends from single feature map $X_{i}$ (i.e., DMSA~\cite{zhu2020deformable}) to multiple feature maps $\bm{X}$=$\left\lbrace X_{i-k}, ..., X_{i}\right\rbrace $ in the temporal domain (see Fig.~\ref{fig:figure6}). Meanwhile, TDMSA exists a total $M$ attention heads for each temporal deformable attention operation, and it can be expressed as:
\begin{equation}\label{tdmsa}
\begin{aligned}
\text{TDMSA}\left(X_{q}, \bm{X}, P_{q}\right) = & \sum_{m=1}^{M} \boldsymbol{W}_{m}\left[  \sum_{l=1}^{L} \sum_{j=i-k}^{i-1} A_{m l j q} \cdot \right.\\
&\left. \boldsymbol{W}_{m}^{\prime} {X}_{j} (P_{jq}+\Delta P_{mljq})\right],
\end{aligned}
\end{equation}
where $P_{jq}$ is a 2D reference point in $j^{th}$ feature map $X_{j}$. $A_{mljq}$ and $\Delta P_{mljq}$ are the learnable attention weight and the sampling offset of the $l^{th}$ sampling point from the $j^{th}$ feature map and the $m^{th}$ attention head.

TDTE includes multiple consecutive stacked blocks (see Fig.~\ref{fig:figure5}). Each block performs MSA, TDMSA, and feed-forward network (FFN) operations. Besides, each operation is followed by the dropout~\cite{srivastava2014dropout} and the layer normalization~\cite{xu2019understanding}. Specifically, TDTE first takes $X_{i}$ as both the query and the key-value via MSA in the timestamp $t_{i}$. Then, TDMSA integrates the current feature map $X_{i}$ and adjacent feature maps $\left\lbrace X_{i-k}, ..., X_{i-1}\right\rbrace $ to leverage rich temporal cues. Finally, FFN is used to output a spatiotemporal representation, shown as $X_I$ or $X_S$ in Fig.~\ref{fig:figure5}. Thus, our TDTE can be formulated as:
\begin{align}
\begin{split}
&\hat{Z_{i}}=\text{MSA}(X_{i}, X_{i}), \\
&\tilde{Z_{i}}=\text{TDMSA}(\hat{Z_{i}}, \left\lbrace X_{i-k}, ..., X_{i-1}\right\rbrace, P_{q} ),  \\
& X_{S}= \text{FFN}(\tilde{Z_{i}}), \\
\end{split}
\end{align}
where $\hat{Z_{i}}$ and $\tilde{Z_{i}}$ are the outputs of MSA and TDMSA operations in our TDTE, respectively. Note that, though frame flow and event flow are processed the same way in this step, there are two TDTEs that do not share weights processing frame flow and event flow separately.

\subsubsection{Temporal Deformable Transformer Decoder}
\label{temporal_decoder}

The goal of our TDTD is to output the final result via a sequence prediction problem. The inputs of TDTD contain both fused feature maps from our fusion module (Section~\ref{fusion}) and object queries. The object queries are a small fixed number (denoted as $N_q$) of learnable positional embeddings. They serve as the "query” part of the input to the TDTD. Each stacked block in the decoder consists of three operations (i.e., MSA, DMSA, and FFN). In the MSA, $N_{q}$ object queries interact with each other in parallel via multi-head self-attention at each decoder layer, where the query elements and key-value elements are both the object queries. As illustrated in Fig.~\ref{fig:figure5}, the DMSA~\cite{zhu2020deformable} transforms the query flows and the fused features from two streams into an output embedding. Ideally, each object query learns to focus on a specific region in images and extracts region features for final detection in this process, similar to the anchors in Faster R-CNN~\cite{ren2017fas} but is simpler and more flexible. Finally, the FFN independently processes each output embedding to predict $N_{q}$ bounding boxes $\bm{B}$=$\left\lbrace B_1, B_2, \cdots, B_{N_q}\right\rbrace$. Notably, $N_{q}$ is typically greater than the actual number of objects in the current scene, so the predicted bounding boxes can either be a detection or a "no object" (labeled as $\phi$) as DETR~\cite{carion2020end} does. In particular, our TDTD achieves more efficient and faster converging by replacing MSA~\cite{carion2020end} with DMSA.

\subsection{Asynchronous Attention-Based Fusion Module}
\label{fusion}

To integrate two heterogeneous visual streams, the most existing fusion strategies~\cite{liu2016combined, jiang2019mixed, li2019event, cao2021fusion, liu2021attention} usually first split asynchronous events into discrete temporal bins synchronized with frames, and then fuse two streams via the post-processing or feature aggregation (e.g., averaging and concatenation). However, these synchronized fusion strategies may have two limitations: (i) The post-processing or feature aggregation operations cannot distinguish the degradation degree of different modalities and regions, which means they are difficult to eliminate the unimodal degradation thoroughly and thus incur the bottleneck of performance improvement; (ii) The joint output frequency may be limited by the sampling rate of conventional frames (e.g., 25 Hz), which fails to meet the needs of fast object detection in real-time high-speed scenarios. Therefore, we design an asynchronous attention-based fusion module to overcome the two limitations mentioned above.

\begin{figure}
    \centerline{\includegraphics[width=\linewidth]{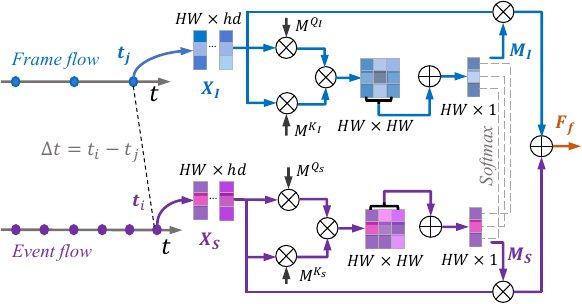}}
	\caption{Overview of the proposed asynchronous attention-based fusion module. It can fuse two heterogeneous data at different sampling timestamps and exploit the high temporal resolution of event streams.}
	\label{fig:figure7}
	\vspace{-0.20cm}
\end{figure}

One key question of two heterogeneous visual streams is how to achieve the high joint output frequency on demand for a given task. In general, the high temporal resolution event stream can be split into discrete event temporal bins, the frequency of which can be flexibly adjusted to be much larger than the conventional frame rate in real-time high-speed scenarios. As shown in Fig.~\ref{fig:figure7}, the event flow has more temporal sampling timestamps than the frame flow in an asynchronous manner. For the timestamp $t_{i}$, $X_{S}$ is the output of the FFN in the TDTE. The proposed fusion strategy first searches the most recent sampling timestamp $t_{j}$ in the frame flow, which can be described as:
\begin{eqnarray}
	j=\mathop{\arg\min}_{k} \left|   t_{i}  - t_{k}  \right|, \Delta t \ge 0,
\end{eqnarray}
where $ \Delta t = t_{i} - t_{k}$ is the temporal difference between the event sampling timestamp and the frame sampling timestamp.

Another key question of two complementary data is how to tackle the unimodal degradation that the feature aggregation operations (e.g., averaging) fail to overcome and implement pixel-wise adaptive fusion for better performance. Unlike the previous weight mask by a multi-layer perceptron (MLP)~\cite{mazhar2021gem}, we compute the pixel-wise weight mask using the attention mechanism, which simplifies our model and endows it with stronger interpretability. Our fusion strategy starts with two intuitive assumptions: (i) the stronger the association between two pixels is, the greater the attention weight between them will be, and (ii) the pixels in degraded regions are supposed to have a weaker association with other pixels. Combining the two points above, we conclude that the sum of attention weights between a pixel and all other pixels represents the degradation degree of that pixel to some extent, and thus can serve as the weight for that pixel during fusion. Next, we take event flow as an example to illustrate how our attention weight mask module (see Fig.~\ref{fig:figure5}) obtains fusion weights. To implement this, we first remove the softmax operation in the standard self-attention to calculate the attention weights $\boldsymbol{W}_{XY}^{S}\in\mathbb{R}^{HW \times HW}$ between each pixel, and accumulate attention weights for each pixel $x$ as $\boldsymbol{W}_{x}^{S}=\sum_{y} \boldsymbol{W}_{xy}^{S}$, which can be summarized as follows:
\begin{equation}
\boldsymbol{W}_{X}^{S} =  \sum_{y} \frac{ ( X_{S} \times \boldsymbol{M}^{Q_{S} }) \times (X_{S} \times \boldsymbol{M}^{K_{S}} )^{T}} {\sqrt{d_k}},
\end{equation}
where $\boldsymbol{W}_{X}^{S} \in\mathbb{R}^{WH \times 1}$ is the weight mask with the width $W$ and the height $H$ for event flow. $\boldsymbol{M}^{Q_{S}}$ and $\boldsymbol{M}^{K_{S}}$ are learnable matrix parameters. $d_k$ is a scaling factor in the self-attention~\cite{vaswani2017attention}. As shown in Fig.~\ref{fig:figure7}, the aforementioned steps are exactly the same for the process of frame flow.

Then, we concatenate the weight masks along the pixel dimension and adopt the softmax to that dimension (i.e., softmax of [$\boldsymbol{W}_{j}^{I}$, $ \boldsymbol{W}_{j}^{S}$], $j = 1, 2, \cdots, HW$) to normalize the pixel-wise attention weights at each pixel. Meanwhile, we further split the weight masks along the pixel dimension to obtain the frame weight masks $\boldsymbol{M}_{I}$ and the event weight masks $\boldsymbol{M}_{S}$ as follows:
\begin{equation}
    \boldsymbol{M}_{I}, \boldsymbol{M}_{S} = \text{split}(\text{softmax}(\left [ \begin{array}{cc}
         \boldsymbol{W}_{1}^{I} & \boldsymbol{W}_{1}^{S} \\
         \vdots & \vdots \\
         \boldsymbol{W}_{HW}^{I} & \boldsymbol{W}_{HW}^{S}
    \end{array}\right ])).
\end{equation}

Finally, we can obtain the fused feature map $F_{f}$ by summing up the unimodal feature maps with the corresponding pixel-wise weight masks by:
\begin{equation}
    F_{f} = (\boldsymbol{M}_{I} \odot X_{I}) \oplus ( \boldsymbol{M}_{S} \odot X_{S}),
\end{equation}
where $\oplus$ denotes the sum operation, and $\odot$ refers to the element-wise multiply operation.

\section{Experiments}
\label{experiment}

\begin{table*}[htbp]
\caption{Performance evaluation of our PKU-DAVIS-SOD dataset in various scenarios. Our SODFormer integrates frames and events to detect objects in a continuous manner, and our baseline processes the single-modality (i.e., frames or events) without utilizing the asynchronous attention-based fusion module.}
\label{tab:table3}
\vspace{-0.30cm}
\begin{center}
	\setlength{\tabcolsep}{2.70mm}{
		\begin{tabular}{l c ccc c ccc cc}
			\hline
			\multirow{2}*{Scenario} & \multirow{2}*{Modality} &\multicolumn{3}{c}{AP$_{50}$} &\multirow{2}*{mAP} &\multirow{2}*{mAP$_{50}$} &\multirow{2}*{mAP$_{75}$} &\multirow{2}*{mAP$_{\text{S}}$} &\multirow{2}*{mAP$_{\text{M}}$} &\multirow{2}*{mAP$_{\text{L}}$}
			\\ \cline{3-5} & & Car & Pedestrian & Two-wheeler &  \\
			\toprule
			\multirow{3}*{Normal} & Events & 0.440 & 0.220 & 0.451 & 0.147 & \textbf{0.371} & 0.090 & 0.072 & 0.268 & 0.526 \\
			& Frames & 0.747 & 0.365 & 0.558 & 0.228 & \textbf{0.557} & 0.138 & 0.166 & 0.336 & 0.539\\
			& Frames + Events & 0.752 & 0.359 & 0.596 & 0.241 & \textbf{0.569} & 0.163 & 0.166 & 0.363 & 0.609 \\
			\hline
			\multirow{3}*{Motion blur} & Events & 0.327 & 0.159 & 0.380 & 0.113 & \textbf{0.289} & 0.064 & 0.051 & 0.181 & 0.255 \\
			& Frames & 0.561 & 0.303 & 0.394 & 0.163 & \textbf{0.419} & 0.096 & 0.100 & 0.201 & 0.365\\
			& Frames + Events & 0.570 & 0.285 & 0.441 & 0.183 & \textbf{0.432} & 0.125 & 0.109 & 0.230 & 0.387\\ 
			\hline
			\multirow{3}*{Low-light} & Events & 0.524 & 0.0002 & 0.294 & 0.093 & \textbf{0.273} & 0.039 & 0.075 & 0.183 & 0.286 \\
			& Frames & 0.570 & 0.128 & 0.357 & 0.114 & \textbf{0.351} & 0.048 & 0.082 & 0.198 & 0.344\\
			& Frames + Events & 0.595 & 0.130 & 0.399 & 0.122 & \textbf{0.374} & 0.048 & 0.091 & 0.206 & 0.376 \\
			\hline
			\multirow{3}*{All} & Events & 0.424 & 0.188 & 0.390 & 0.128 & \textbf{0.334} & 0.071 & 0.065 & 0.210 & 0.348 \\
			& Frames & 0.700 & 0.316 & 0.452 & 0.195 & \textbf{0.489} & 0.116 & 0.142 & 0.264 & 0.417\\
			& Frames + Events & 0.705 & 0.313 & 0.493 & 0.207 & \textbf{0.504} & 0.133 & 0.144 & 0.285 & 0.454\\
			\hline
	\end{tabular}}
\end{center}
\vspace{-0.10cm}
\end{table*}

\begin{figure*}[t]
	\begin{subfigure}[b]{0.333\linewidth}
		\centering
		\centerline{\includegraphics[height=8.92cm]{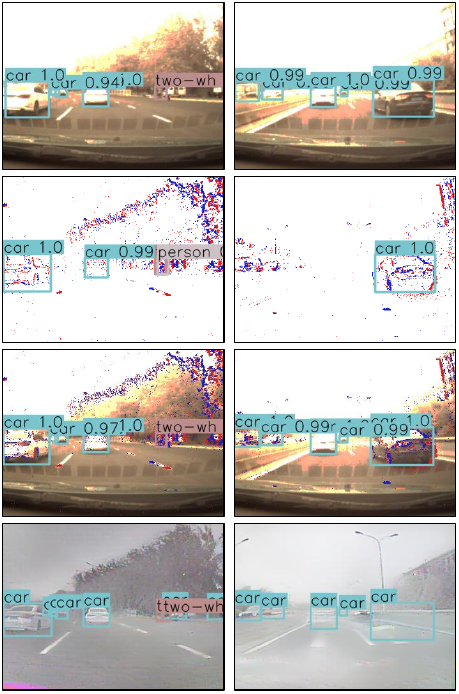}}
		\caption{Normal}
		\label{fig:8(a)}
	\end{subfigure}
	\begin{subfigure}[b]{0.333\linewidth}
		\centering
		\centerline{\includegraphics[height=8.92cm]{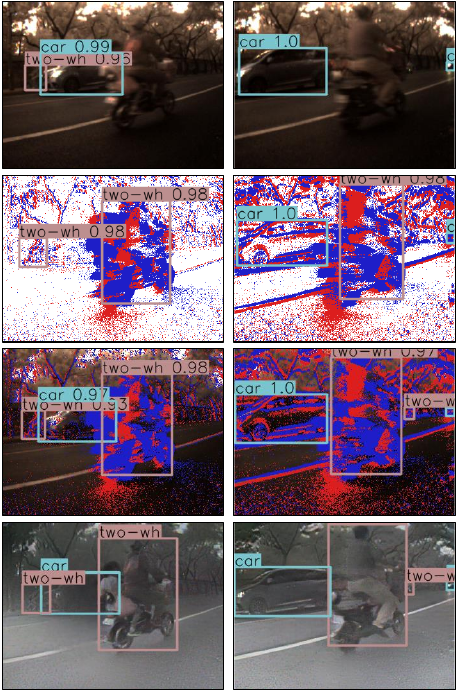}}
		\caption{Motion blur}
		\label{fig:8(b)}
	\end{subfigure}
	\begin{subfigure}[b]{0.333\linewidth}
		\centering
		\centerline{\includegraphics[height=8.92cm]{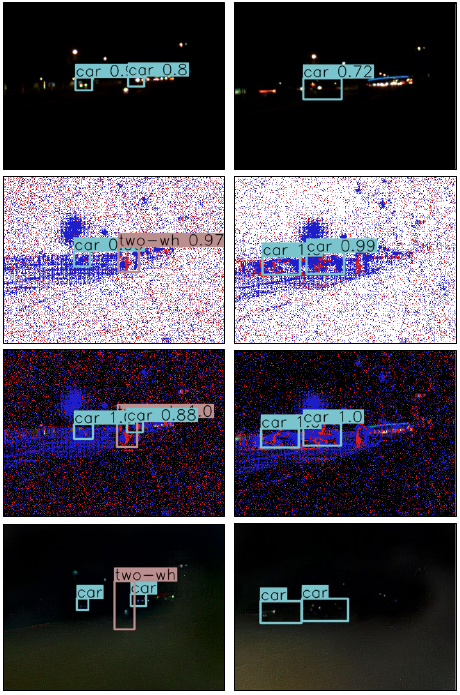}}
		\caption{Low-light}
		\label{fig:8(c)}
	\end{subfigure}
	\vspace{-0.46cm}
	\caption{Representative visualization results on various scenarios of our PKU-DAVIS-SOD dataset. The four rows from top to bottom refer to our baseline using RGB frames, our baseline using event images, our SODFormer using RGB frames and DVS events, and ground truth labeled in event reconstructed images~\cite{rebecq2019events}.}
	\label{fig:figure8}
	\vspace{-0.35cm}
\end{figure*}

This section first describes the experimental setting and implementation details (Section~\ref{setting}). Then, the effective test (Section~\ref{effective_test}) is conducted to verify the validity of our SODFormer, which contains a performance evaluation in various scenarios and a comparison to related detectors. Meanwhile, we further implement the ablation study (Section~\ref{ablation_test}) to see why and how our approach works, where we investigate the influences of each designed module and parameter settings. Finally, the scalability test (Section~\ref{scalability_test}) provides interpretable explorations of our SODFormer.

\subsection{Experimental Settings} \label{setting}

In this part, we will present the dataset, implementation details, and evaluation metrics as follows. 

\emph{Dataset and Setup}. Our PKU-DAVIS-SOD dataset is designed for streaming object detection in driving scenes, which provides two heterogeneous and temporally long-term visual streams with manual labels at 25 Hz (see Section~\ref{dataset}). This newly built dataset contains three subsets including 671.3k labels for training, 194.7k for validation, and 214.1k for testing. Each subset is further split into three typical scenarios (i.e., normal, low-light, and motion blur). 

\emph{Implementation Details}. We select event images~\cite{maqueda2018event} as the event representation and the ResNet50~\cite{he2016deep} as the backbone to achieve a trade-off between accuracy and speed. All parameters of the backbone and the spatial Transformer encoder (STE) are shared among different temporal bins of the same modality. We set the attention head $M$ to 8 and the sampling point $L$ to 4 for the deformable multi-head self-attention (DMSA) in Eq.(\ref{dmsa}) and the temporal deformable multi-head self-attention (TDMSA) in Eq.(\ref{tdmsa}). The temporal aggregation size in TDMSA is set to 9 owning to the balance of the accuracy and the speed. During training, the height of an image is randomly resized to a number from [256: 576: 32], and the width is obtained accordingly while maintaining the aspect ratio. Similarly, for evaluation, all images are resized to 352 in height and 468 in width to maintain the aspect ratio. Other data augmentation methods such as crop and horizontal flip are not utilized because our temporal Transformer requires accurate reference points. Following the Deformable DETR~\cite{zhu2020deformable}, we adopt the matching cost and the Hungarian loss for training with loss weights of 2, 5, 2 for classification, $L_1$ and GIoU, respectively. All networks are trained for 25 epochs using the Adam optimizer~\cite{kingma2014adam}, which is set with the initial learning rate of $2 \times 10^{-4}$, the decayed factor of 0.1 after the 20th epoch, and the weight decay of $10^{-4}$. All experiments are conducted on NVIDIA Tesla V100 GPUs.

\emph{Evaluation Metrics}. To compare different approaches, the mean average precision (e.g., COCO mAP~\cite{lin2014microsoft}) and running time ($ms$) are selected as two evaluation metrics, which are the most broadly utilized in the object detection task. In the effective test, we give a comprehensive evaluation using average precision with various IoUs (e.g., AP, AP$_{0.5}$, and AP$_{0.75}$) and AP across different scales (i.e., AP$_\text{S}$, AP$_\text{M}$, and AP$_\text{L}$). In other tests, we report the detection performance using the AP$_{0.5}$. Following the video object detection, we compute the AP for each class and the mAP for three classes in all labeled timestamps. In other words, we evaluate the object detection performance with the output frequency of 25 Hz in every labeled timestamp.

\begin{table*}[htbp]
	\caption{Comparison with state-of-the-art methods and our baselines on our PKU-DAVIS-SOD dataset. Our SODFormer, making complementary use of RGB frames and DVS events, outperforms seven state-of-the-art methods involving event cameras, four using RGB frames, and two joint object detectors. * denotes that a method leverages temporal cues.}
	\label{tab:table4}
	\vspace{-0.40cm}
	\begin{center}
		\setlength{\tabcolsep}{1.10mm}{
			\begin{tabular}{l c ccc cc}
				\hline
				Modality & Method &Input representation &Backbone &Temporal  & mAP$_{50}$ & Runtime ($ms$) \\
				\toprule
				\multirow{8}*{Events} & SSD-events~\cite{iacono2018towards} & Event image & SSD & No & 0.221 & 7.2  \\
				& NGA-events~\cite{hu2020learning} & Voxel grid & YOLOv3 & No & 0.232  & 8.0   \\
				& YOLOv3-RGB~\cite{redmon2018yolov3} & Reconstructed image & YOLOv3 & No & 0.244 & 178.51 \\
                & Faster R-CNN~\cite{ren2017fas} & Event image & R-CNN & No & 0.251 & 74.5\\
				& Deformable DETR~\cite{zhu2020deformable} & Event image & DETR & No & 0.307 & 21.6 \\
				& LSTM-SSD*~\cite{zhu2018mobile} & Event image & SSD & Yes & 0.273 & 22.7 \\
                & ASTMNet*~\cite{li2022asynchronous} & Event embedding & Rec-Conv-SSD & Yes & 0.291 & 21.3 \\
				& \textbf{Our baseline}* & Event image & Our spatiotemporal Transformer & Yes & 0.334 & 25.0 \\
				\hline
				\multirow{5}*{Frames} & Faster R-CNN~\cite{ren2017fas} & RGB frame & R-CNN & No & 0.443 & 75.2\\
                & YOLOv3-RGB~\cite{redmon2018yolov3} & RGB frame & YOLOv3 & No & 0.426 & 7.9 \\
				& Deformable DETR~\cite{zhu2020deformable} & RGB frame & DETR & No & 0.461 & 21.5\\
				& LSTM-SSD*~\cite{zhu2018mobile} & RGB frame & SSD & Yes & 0.456 & 22.4 \\
				& \textbf{Our baseline}* & RGB frame & Our spatiotemporal Transformer & Yes & 0.489 & 24.9 \\
				\hline
				\multirow{3}*{Events + Frames} & MFEPD~\cite{jiang2019mixed} & Event image + RGB frame & YOLOv3 & No & 0.438 & 8.2\\
				& JDF~\cite{li2019event} & Channel image + RGB frame & YOLOv3 & No & 0.442 & 8.3\\
				& \textbf{Our SODFormer} & Event image + RGB frame & Deformable DETR & Yes & 0.504 & 39.7\\
				\hline
		\end{tabular}}
	\end{center}
	\vspace{-0.30cm}
\end{table*}

\subsection{Effective Test} \label{effective_test}

The objective of this part is to assess the validity of our SODFormer, so we implement two main experiments including performance evaluation in various scenarios (Section~\ref{evaluation_each_scenario}) and comparison with state-of-the-art methods (Section~\ref{state_of_the_art}) as follows.

\subsubsection{Performance Evaluation in Various Scenarios} \label{evaluation_each_scenario}

To give a comprehensive evaluation of our PKU-DAVIS-SOD dataset, we report the quantization results (see Table~\ref{tab:table3}) and the representative visualization results (see Fig.~\ref{fig:figure8}) from three following perspectives.

\emph{Normal Scenarios}. Performance evaluations of all modalities (i.e., frames, events, and two visual streams) in normal scenarios can be found in Table~\ref{tab:table3}. Our single-modality baseline only processes frames or events without using the asynchronous attention-based fusion module. We can see that our baseline using RGB frames achieves better performance than using events in normal scenarios. This is because the event camera has a flaw with weak texture in the spatial domain, thus only dynamic events (i.e., brightness changes) are hard to achieve high-precision recognition, especially in static or extremely slow motion scenes (see Fig.~\ref{fig:figure8}(\subref{fig:8(a)})). On the contrary, RGB frames can provide static fine textures (i.e., absolute brightness) under ideal conditions. In particular, our SODFormer obtains the performance enhancement from the single stream to two visual streams via introducing the asynchronous attention-based fusion module.

\emph{Motion Blur Scenarios}. By comparing normal scenarios and motion blur scenarios in Table~\ref{tab:table3}, we can find that the detection performance using events degrades less than that utilizing frames. This can be explained by that RGB frames appear as motion blur in high-speed motion scenes (see Fig.~\ref{fig:figure8}(\subref{fig:8(b)})), resulting in a remarkable decrease in object detection performance. Note that, by introducing the event stream, the performance of our SODFormer can appreciably enhance over our baseline using RGB frames.

\emph{Low-Light Scenarios}. As illustrated in Table~\ref{tab:table3}, the performance of our baseline using RGB frames drops sharply by 0.206 in low-light scenarios. Meanwhile, although the mAP of DVS events also drops by 0.098 in absolute term, it is significantly less compared to that of RGB frames (i.e., 0.098 vs. 0.206). On a relative basis, there is also an evident difference as the mAP of RGB frames and DVS events drop by 0.370 and 0.264, respectively. This may be caused by the fact that the event camera has the advantage of HDR to sense moving objects in extreme light conditions (see Fig.~\ref{fig:figure8}(\subref{fig:8(c)})). Notably, only the AP$_{50}$ of "Car" objects increases while the others decrease. We attribute this to the different image quality of different classes of objects in the event modality. Intuitively, "Car" objects tend to have more distinct outlines and higher image quality. As a result, "Car" objects are less distributed by the increase of noise in low-light scenes~\cite{guo2023low} and thus achieve comparable performance to that in normal scenes. More specifically, our SODFormer outperforms the baseline using RGB frames by a large margin (0.374 versus 0.351) in low-light scenarios.

In summary, our SODFormer consistently obtains better performance than the single-modality methods in three scenarios meanwhile keeping relatively comparable computational speed. By introducing DVS events, the mAP on our PKU-DAVIS-SOD dataset improves an average of 1.5\% over using RGB frames. What's more, representative visualization results in Fig.~\ref{fig:figure8} show that RGB frames fail to detect objects in high-speed motion blur and low-light scenarios. However, the event camera, offering high temporal resolution and HDR, provides new insight into addressing the shortages of conventional cameras.

\begin{figure*}
	\centering
	\centerline{\includegraphics[width=\linewidth]{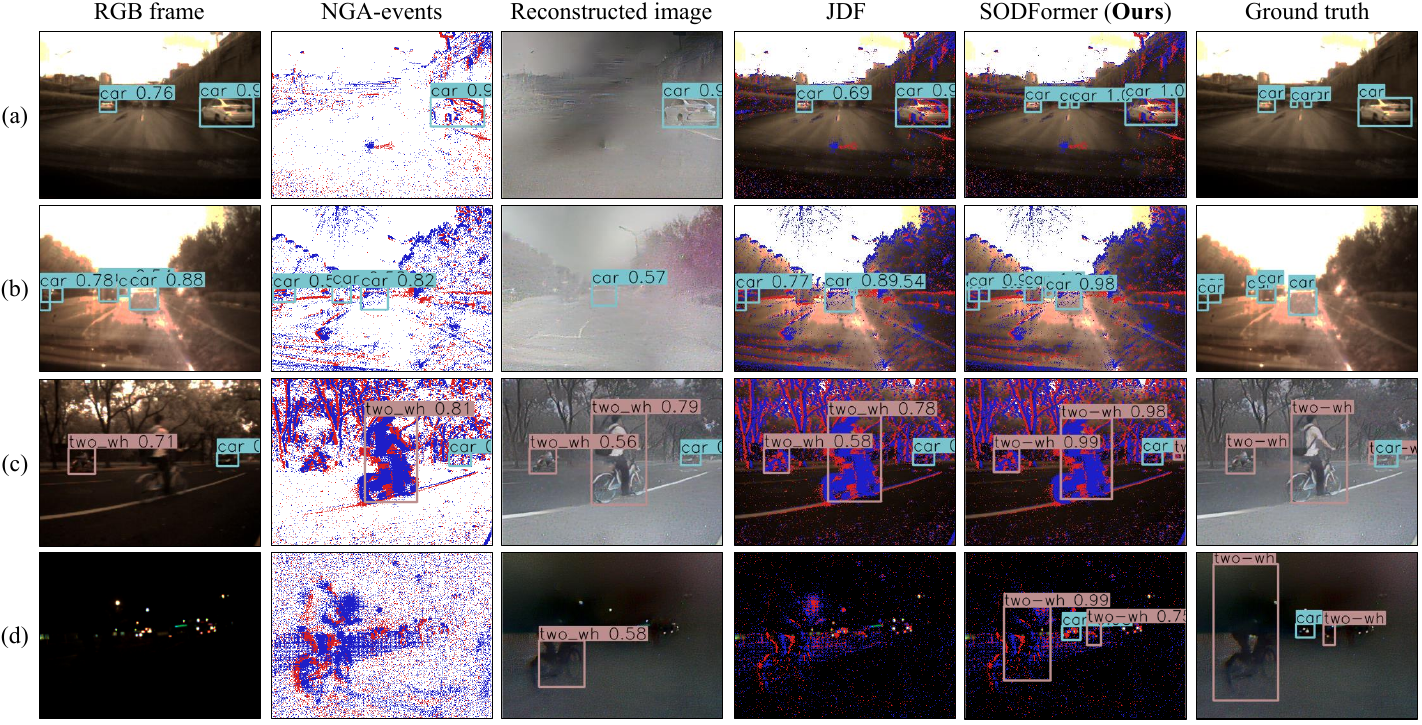}}
	\caption{Representative examples of different object detection results on our PKU-DAVIS-SOD dataset. (a) Moving cars in normal scenario. (b) Rushing cars in overexposure scenario. (c) High-speed running two-wheelers with motion blur. (d) Traveling two-wheelers in low-light scenario.}
	\label{fig:figure9}
	\vspace{-0.20cm}
\end{figure*}

\subsubsection{Comparison with State-of-the-Art Methods} \label{state_of_the_art}

We will instigate why and how our SODFormer works from the following three perspectives.

\emph{Evaluation on DVS Modality}. To evaluate our temporal Transformer for DVS events, we compare our baseline* (i.e., our SODFormer without the asynchronous attention-based fusion module) with five feed-forward event-based object detectors (i.e., event image for SSD~\cite{iacono2018towards}, voxel grid for YOLOv3~\cite{hu2020learning}, reconstructed image for YOLOv3~\cite{redmon2018yolov3}, event image for Faster R-CNN~\cite{ren2017fas}, and event image for Deformable DETR~\cite{zhu2020deformable}) and two recurrent object detectors (event image for LSTM-SSD~\cite{zhu2018mobile} and event embedding for ASTMNet~\cite{li2022asynchronous}) that leverages temporal cues. As shown in Table~\ref{tab:table4}, our baseline*, utilizing event images for our temporal Transformer, achieves significant improvement over all these object detectors. Besides, our other baseline, using E2VID~\cite{rebecq2019events} to reconstruct frames, performs better than other input representations (e.g., event image and voxel grid), but a long time is spent on the first stage of image reconstruction and then on the second stage of object detection.


\emph{Evaluation on RGB Modality}. We present four state-of-the-art object detectors for RGB frames including (i) Single RGB frame for Faster R-CNN, (ii) Single RGB frame for YOLOv3, (iii) Single RGB frame for Deformable DETR, (iv) Sequential RGB frames for LSTM-SSD to compare them with our baseline* that utilizes sequential RGB frames for our spatiotemporal Transformer. Compared to three object detectors utilizing single RGB frames, our baseline* obtains the best performance via introducing the temporal Transformer for RGB frames (see Table~\ref{tab:table4}). Furthermore, it obtains better performance than LSTM-SSD while maintaining comparable computational speed. This is because the designed spatiotemporal Transformer is more effective than the standard convolutional-recurrent operation. In other words, the results agree with our motivation that leveraging rich temporal cues for streaming object detection is beneficial.

\emph{Benefit From Multimodal Fusion}. From Table~\ref{tab:table4}, we make a comparison of our SODFormer between two existing joint object detectors (e.g., MFEPD~\cite{jiang2019mixed} and JDF~\cite{li2019event}). Apparently, our SODFormer outperforms two competitors by a large margin. We can find that our SODFormer, incorporating the temporal Transformer and the asynchronous attention-based fusion module, outperforms four state-of-the-art detectors and our eight baselines. More precisely, by introducing DVS events, our SODFormer has a 1.5\% improvement over the best baseline using sequential RGB frames. Compared with the best competitor using DVS events, our SODFormer truly shines (0.504 versus 0.334), which indicates that RGB frames with fine textures may be very significant for high-precision object detection.

We further present some visualization results on our PKU-DAVIS-SOD dataset in Fig.~\ref{fig:figure9}. Note that, our SODFormer outperforms the single-modality methods and the best multi-modality competitor (i.e., JDF~\cite{li2019event}). Unfortunately, RGB frames fail to detect objects in high-speed or low-light scenes. Much to our surprise, our SODFormer can perform robust detect objects in challenging situations, due to its high temporal resolution and HDR properties that are inherited from DVS events, as well as its fine texture that comes from RGB frames.

\subsection{Ablation Test} \label{ablation_test}

In this section, we implement an ablation study to investigate how parameter setting and each design choice influence our SODFormer.

\begin{figure*}[t]
	\centering
	\centerline{\includegraphics[width=\linewidth]{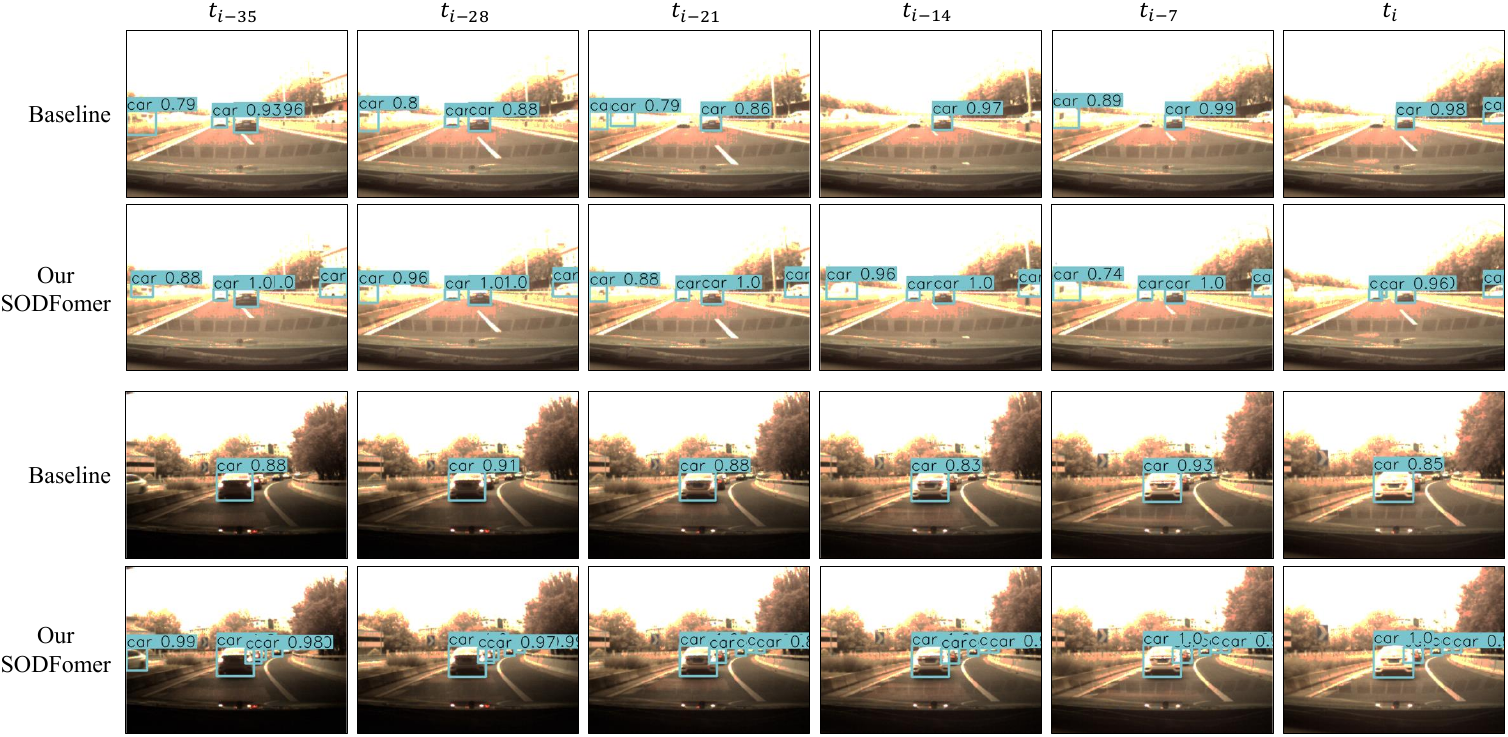}}
	\caption{Representative visualization results in continuous sequences of our PKU-DAVIS-SOD dataset. Note that, our SODFormer achieves better object detection performance than the feed-forward baseline without using temporal cues, especially when involving small objects in the distance or partially occluded objects.}
	\label{fig:figure10}
	\vspace{-0.20cm}
\end{figure*}


\subsubsection{Contribution of Each Component} \label{contribution}

\begin{table}
	\caption{The contribution of each component to our SODFormer on our PKU-DAVIS-SOD dataset. All results are obtained with the baseline using RGB frames for Deformable DETR~\cite{zhu2020deformable}.}
	\label{tab:table5}
	\vspace{-0.40cm}
	\begin{center}
		\setlength{\tabcolsep}{2.40mm}{
		\small
			\begin{tabular}{l cccc}
				\hline
				Method & Baseline & (a) & (b) & \textbf{Ours} \\
				\toprule
				Events &  &  & \ding{51} & \ding{51} \\
				Temporal transformer &  & \ding{51} & \ding{51} & \ding{51}\\
				Fusion Module &  &  &  & \ding{51} \\
				\hline
				mAP$_{50}$ & 0.461 & 0.489 & 0.494 & \textbf{0.504}\\
				Runtime ($ms$) & 21.5 & 24.9 & 39.4 & 39.7\\
				\hline
		\end{tabular}}
	\end{center}
	\vspace{-0.30cm}
\end{table}

To explore the impact of each component on the final performance, we choose the feed-forward detector (i.e., using RGB frame for Deformable DETR~\cite{zhu2020deformable}) as the baseline. As illustrated in Table~\ref{tab:table5}, three methods, namely (a), (b), and our SODFormer, utilize temporal Transformer, averaging fusion module and asynchronous attention-based fusion module respectively, consistently achieve higher performance on our PKU-DAVIS-SOD dataset than the baseline. More specifically, method (a), adopting the temporal Transformer to leverage rich temporal cues, obtains a 2.8\% mAP improvement over the baseline. Comparing method (b) with (a), the absolute promotion is merely 0.5\%, which indicates that it is insufficient to combine RGB frames and DVS events using the averaging operation. Moreover, our SODFormer, adopting the asynchronous attention-based fusion strategy to replace the averaging fusion, achieves a 1.0\% mAP improvement over method (b) while maintaining comparable computational complexity. Intuitively, our SODFormer employs these effective components to process two heterogeneous visual streams and achieves robust object detection on our PKU-DAVIS-SOD dataset.

\subsubsection{Influence of Temporal Aggregation Size} \label{aggregation_size}

To analyze the temporal aggregation strategy in our SODFormer, we set the temporal Transformer with different sizes (e.g., 1, 3, 5, 9, and 13) of temporal aggregation. We refer to this as temporal aggregation size, which represents the number of event temporal bins $S$ and adjacent frames $I$ to process at a given timestamp $t_i$. It fits in Eq.~\ref{tdmsa} as $k+1$. As depicted in Table~\ref{tab:table6}, we find that our strategy improves mAPs by 1.2\%, 1.9\%, 2.8\%, and 2.7\% when compared to a single temporal aggregation size. Note that, by aggregating visual streams in a larger temporal size, richer temporal cues can be utilized in the temporal Transformer, resulting in better object detection performance. Nevertheless, as the temporal aggregation size becomes larger, the computational time also gradually increases. Additionally, we find that the model's capacity for modeling temporally long-term dependency is finite with the increase in temporal aggregation size. In this study, our temporal aggregation size is set to 9 for a trade-off between accuracy and speed.

\begin{table}
	\renewcommand{\arraystretch}{1.00}
	\caption{The influence of temporal aggregation size on our PKU-DAVIS-SOD dataset. The single-modality baseline is conducted using RGB frames for Deformable DETR\cite{zhu2020deformable}.}
	\label{tab:table6}
	\vspace{-0.40cm}
	\begin{center}
		\setlength{\tabcolsep}{2.00mm}{
			\begin{tabular}{l ccccc}
				\hline
				Temporal aggregation size & 1 & 3 & 5 & 9 & 13  \\
				\hline
				mAP$_{50}$ & 0.461 & 0.473 & 0.480 & 0.489 & 0.488\\
				Runtime ($ms$) & 21.5 & 24.1 & 24.2 & 24.9 & 25.5\\
				\hline
		\end{tabular}}
	\end{center}
	\vspace{-0.30cm}
\end{table}

To improve the interpretability of streaming object detection, we further present some comparative visualization results about whether they utilize rich temporal cues or not (see Fig.~\ref{fig:figure10}). The feed-forward baseline using RGB frames suffers from some failure cases involving small objects and occluded objects, as shown in the first and third rows in Fig.~\ref{fig:figure10}. Fortunately, our SODFormer overcomes these limitations via leveraging rich temporal cues from adjacent frames or event streams. Therefore, a trade-off between accuracy and speed may be beneficial and necessary for certain scenarios requiring a higher level of accuracy in object detection.

\begin{table}
	\caption{Comparison of the proposed asynchronous attention-based fusion module with typical fusion strategies on our PKU-DAVIS-SOD dataset.}
	\label{tab:table7}
	\vspace{-0.40cm}
	\begin{center}
		\setlength{\tabcolsep}{5.50mm}{
			\small
			\begin{tabular}{l cc}
				\hline
				Method & mAP$_{50}$ & Runtime ($ms$) \\
				\toprule
				NMS~\cite{hosang2017learning} & 0.438 & 8.2 \\
				Concatenation~\cite{xu2020multi} & 0.478 & 39.6\\
				Averaging~\cite{li2020mdlatlrr} & 0.494 & 39.4 \\
				\textbf{Ours} & 0.504 & 39.7 \\
				\hline
		\end{tabular}}
	\end{center}
	\vspace{-0.30cm}
\end{table}

\begin{figure}
	\centering
	\centerline{\includegraphics[width=\linewidth]{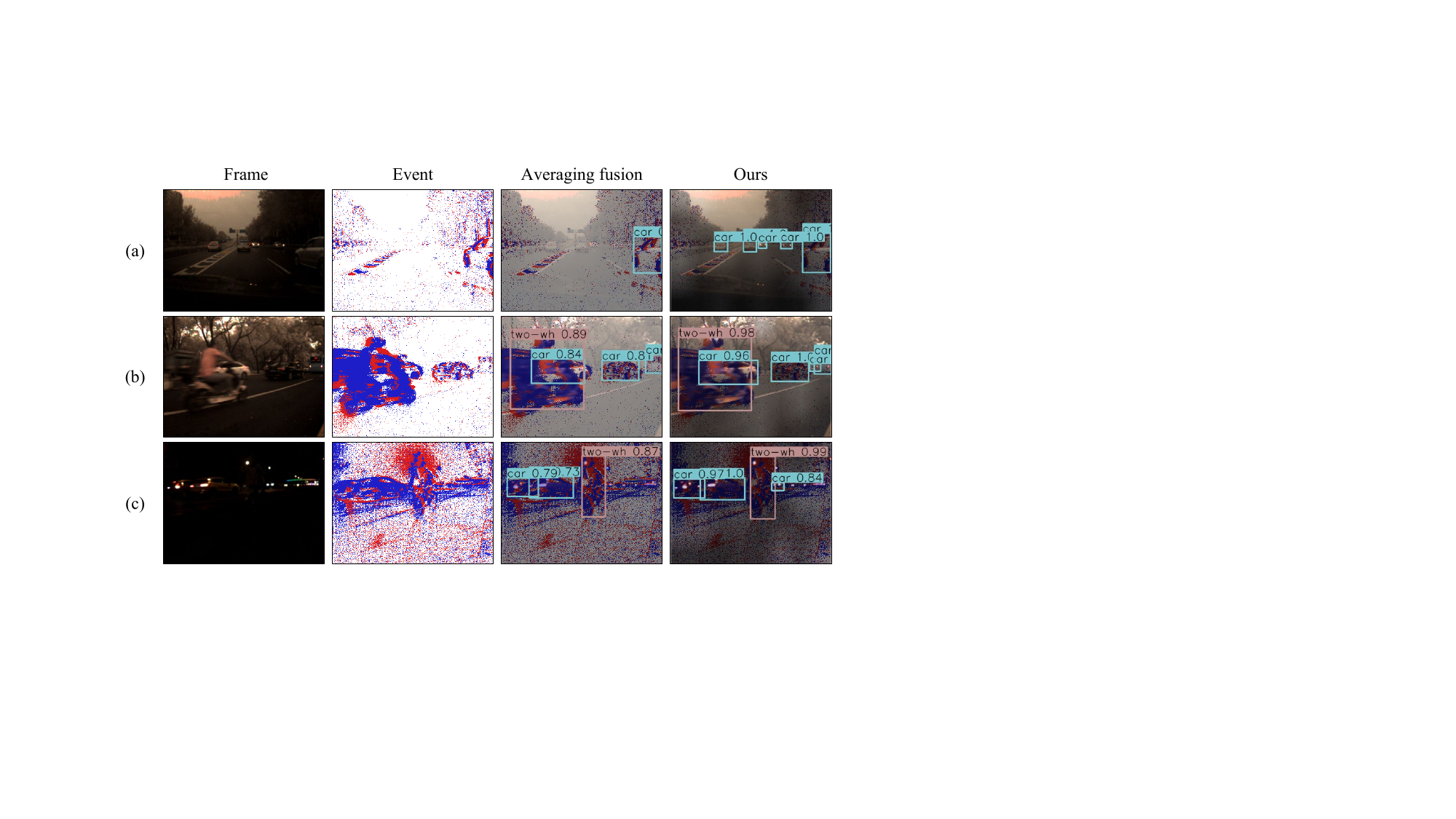}}
	\caption{Representative instances of fusion results on our PKU-DAVIS-SOD dataset. The four columns from left to right are RGB frames, event images, fused images using averaging fusion, and fused images using our fusion strategy.}
	\label{fig:figure11}
	\vspace{-0.30cm}
\end{figure}

\subsubsection{Influence of Fusion Strategy} \label{fusion_strategy}
In order to evaluate the effectiveness of our fusion module, we compare it with some typical fusion methods in Table~\ref{tab:table7}. Specifically, the concatenation and averaging method are both applied to the feature maps produced by the temporal Transformer, referred to as $X_S$ and $X_I$ in Fig.~\ref{fig:figure5}. The $X_I$ and $X_S$ are concatenated along the last dimension for concatenation and are averaged when averaging. The fused feature map is input to the decoder in both methods. Note that, our approach obtains the best performance against the post-processing operations (e.g., NMS~\cite{hosang2017learning}) and end-to-end feature aggregation techniques (e.g., averaging~\cite{li2020mdlatlrr} and concatenation~\cite{xu2020multi}). More precisely, our strategy gets around 6.6\%, 2.6\% and 1.0\% improvements on our PKU-DAVIS-SOD dataset with three corresponding methods. Compared with the averaging operation, our approach improves the mAP from 0.494 to 0.504 while maintaining comparable computation time. Additionally, we present some visualization comparison results in Fig.~\ref{fig:figure11}. Three representative instances show that our approach performs better than the averaging operation. This is because our asynchronous attention-based fusion module can adaptively generate pixel-wise weight masks and eliminate unimodal degradation thoroughly.

\begin{figure*}[htbp]
	\centering
	\centerline{\includegraphics[width=\linewidth]{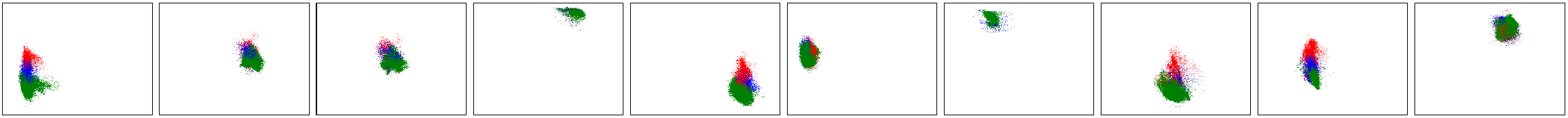}}
	\caption{Visualization of all predicted bounding boxes from the test subset of our PKU-DAVIS-SOD dataset. We present 10 representative examples from 300 predicted slots. Boxes are regarded as points with centers obtained from their coordinates. Blue points refer to large vertical boxes, red points to horizontal boxes, and green points to small boxes. Notably, each slot can be trained to focus on certain areas and object sizes with a variety of operating modes.}
	\label{fig:figure12}
	\vspace{-0.30cm}
\end{figure*}

\begin{table}
	\caption{Comparison of our SODFormer with various event representations on our PKU-DAVIS-SOD dataset.}
	\label{tab:table10}
	\vspace{-0.40cm}
	\begin{center}
		\setlength{\tabcolsep}{3.4mm}{
			\small
			\begin{tabular}{l cc}
				\hline
				Method & mAP$_{50}$ & Runtime ($ms$) \\
				\toprule
				Sigmoid representation~\cite{chen2018pseudo} & 0.309 & 39.4 \\
				Voxel grid~\cite{zhu2019unsupervised} & 0.491 & 41.5\\
				Event images~\cite{maqueda2018event} & 0.504 & 39.7 \\
				\hline
		\end{tabular}}
	\end{center}
	\vspace{-0.20cm}
\end{table}

\subsubsection{Influence of Event Representation}\label{event_representation}

To verify the generality of our SODFormer for various event representations, we compare three typical event representations (i.e., event images~\cite{maqueda2018event}, voxel grids~\cite{zhu2019unsupervised} and sigmoid representation~\cite{chen2018pseudo}) owning to an accuracy-speed trade-off. As illustrated in Table~\ref{tab:table10}, the performance of our SODFormer varies with different event representations, but their running speed is almost identical. It indicates that our SODFormer can provide a generic interface in combination with various input representations, such as, image-like representations (e.g., event images~\cite{maqueda2018event} and sigmoid representation~\cite{chen2018pseudo}) and spatiotemporal representations (i.e., voxel grids~\cite{zhu2019unsupervised}). Meanwhile, it is obvious that the performance is highly dependent on how well the representation is and drops heavily with worse methods (e.g., sigmoid representation). Indeed, we believe that a good event representation makes asynchronous events directly compatible with our SODFormer while maximizing the detection performance.

\subsubsection{Influence of the Number of Transformer's Layer} \label{transformer_layer}
We will analyze the effect of the number of layers in the designed temporal Transformer on the final performance from the following two perspectives.

\emph{The Number of TDTE's Layer}. As shown in Table~\ref{tab:table8}, we first explore the influence of the number of encoder layers in our TDTE on our PKU-DAVIS-SOD dataset. To our surprise, as the number of encoder layers increases, our SODFormer shows only slight improvements up to 6 layers and a rapid decline due to overfitting after that, yet the inference time is gradually getting longer. Considering that increasing the number of encoder layers in our TDTE doesn't improve the detection performance but increases the computational time, the number of encoder layers in our TDTE is set to 6.

\begin{table}
	\renewcommand{\arraystretch}{1.00}
	\caption{The influence of the number of TDTE's layers on our PKU-DAVIS-SOD dataset.}
	\label{tab:table8}
	\vspace{-0.50cm}
	\begin{center}
		\setlength{\tabcolsep}{1.60mm}{
			\begin{tabular}{l ccccc}
				\hline
				The number of encoder layers & 2 & 4 & 6 & 8 & 12\\
				\toprule
				mAP$_{50}$ & 0.481 & 0.480 & 0.489 & 0.471 & 0.450 \\
				Runtime ($ms$) & 23.5 & 24.2 & 24.9 & 25.2 & 26.1 \\
				\hline
		\end{tabular}}
	\end{center}
	\vspace{-0.20cm}
\end{table}

\emph{The Number of TDTD's Layer}. Table~\ref{tab:table9} further illustrates the influence of decoder layers in our TDTD on our PKU-DAVIS-SOD dataset. We can find that the best performance occurs when we set the number of decoder layers to 8. However, the mAP is almost unchanged and the computational time increases rapidly when the number of decoder layers exceeds 6. In order to balance detection performance and computational complexity, it is reasonable to set the number of TDTD's layers to 6.

\begin{table}
	\renewcommand{\arraystretch}{1.00}
	\caption{The influence of the number of TDTD's layers on our PKU-DAVIS-SOD dataset.}
	\label{tab:table9}
	\vspace{-0.40cm}
	\begin{center}
		\setlength{\tabcolsep}{1.60mm}{
			\begin{tabular}{l ccccc}
				\hline
				The number of decoder layers & 2 & 4 & 6 & 8 & 12 \\
				\toprule
				mAP$_{50}$ & 0.476 & 0.480 & 0.489 & 0.491 & 0.489 \\
				Runtime ($ms$) & 21.1 & 22.3 & 24.9 & 29.5 & 34.8\\
				\hline
		\end{tabular}}
	\end{center}
	\vspace{-0.40cm}
\end{table}

\subsection{Scalability Test} \label{scalability_test}
This subsection will present the predicted output slot analysis (Section~\ref{slot_analysis}) and the visualization of the designed temporal deformable attention (Section~\ref{visualization_attention}). Then, we further present how to process two heterogeneous visual streams in an asynchronous manner (Section~\ref{asynchronous_analysis}). Finally, we analyze some failure cases of our SODFormer (Section~\ref{failure_case}).

\begin{figure*}[htbp]
	\centering
	\centerline{\includegraphics[width=\linewidth]{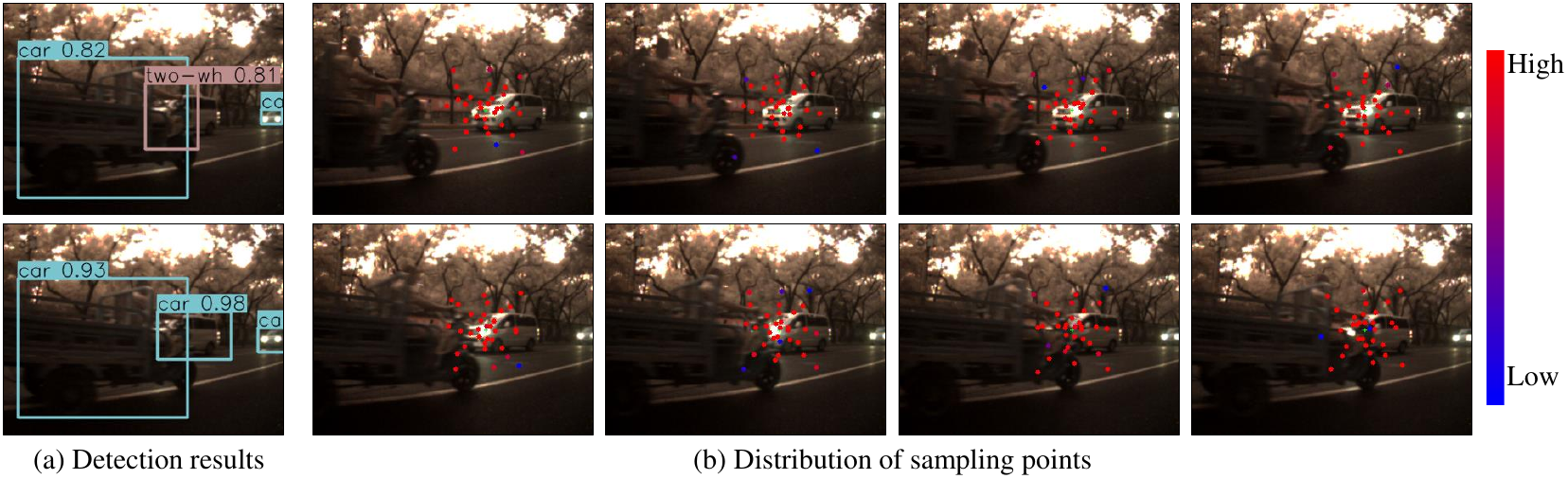}}
	\caption{Visualization of TDMSA in the last layer of our TDTE. (a) Comparison of object detection results between the feed-forward baseline (top) and our unimodal SODFormer without using events (bottom). (b) Distribution of sampling points in reference frames. The reference point is denoted by a green cross marker, and each sampling point is marked as a color-filled circle whose color represents the size of its weight.}
	\label{fig:figure13}
	\vspace{-0.30cm}
\end{figure*}

\begin{figure}[htbp]
	\centering
	\centerline{\includegraphics[width=\linewidth]{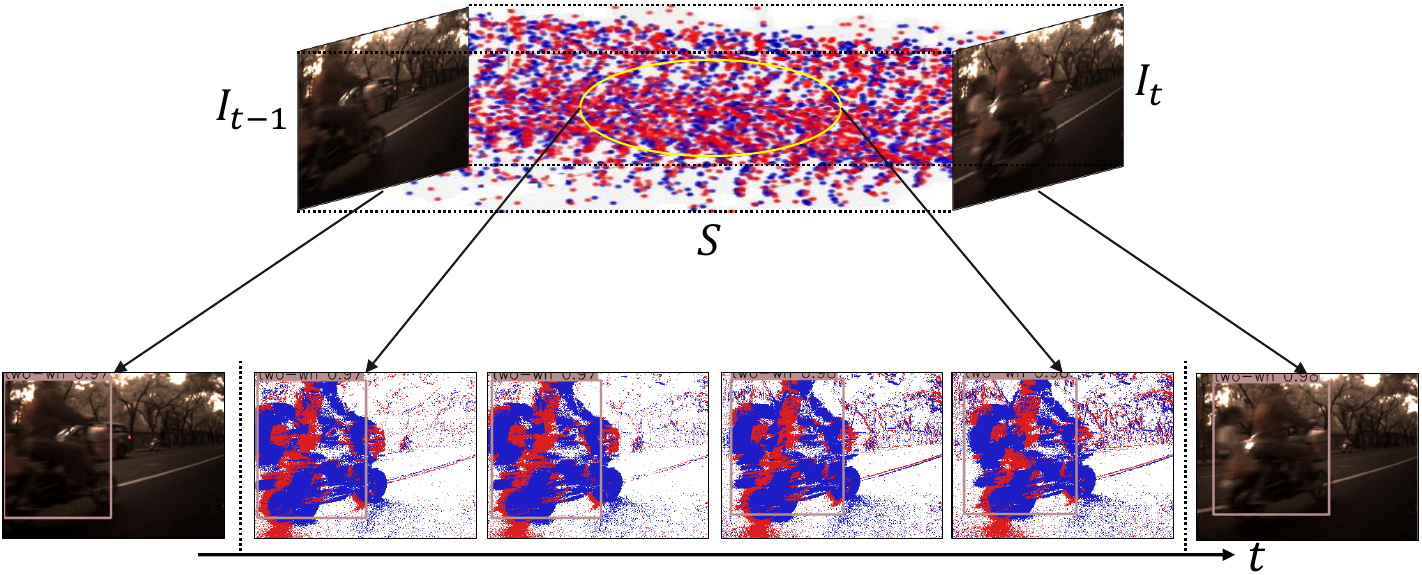}}
	\caption{Visualization of our asynchronous fusion strategy. The top figure depicts two adjacent frames and an event stream. The bottom figures present the detection performance of the two frames and the four sampling timestamps from the continuous event stream.}
	\label{fig:figure14}
	\vspace{-0.30cm}
\end{figure}

\subsubsection{The Predicted Output Slot Analysis} \label{slot_analysis}
To increase the interpretability of our SODFormer, we visualize the predicted bounding boxes of different slots for all labeled timestamps in the test subset of our PKU-DAVIS-SOD dataset. Inspired by DETR~\cite{carion2020end}, each predicted box is represented as a point whose coordinates are the center normalized by image size. Fig.~\ref{fig:figure12} shows 10 representative examples out of 300 predicted slots in our temporal deformable Transformer decoder (TDTD). Apparently, we find that each slot can learn to focus on certain areas and bounding box sizes. Actually, our SODFormer can learn a different specialization for each object query slot, and all slots can cover the distribution of all objects in our PKU-DAVIS-SOD dataset.

\subsubsection{Visualization of Temporal Deformable Attention} \label{visualization_attention}
To better understand the proposed temporal deformable attention, we visualize sampling points and attention weights of the last layer in our temporal deformable Transformer encoder (TDTE). As shown in Fig.~\ref{fig:figure13}(a), our feed-forward baseline fails to detect the middle car owning to the occlusion, while our unimodal SODFormer succeeded by utilizing rich temporal cues in adjacent frames. In addition, we present the distribution of sampling points in the last layer of our TDTE (see Fig.~\ref{fig:figure13}(b)). All sampling points are mapped to the corresponding location in RGB frames, and each sampling point is marked as a color-filled circle whose color represents the size of the weight. The reference point is labeled as a green cross marker. We can find that the distribution of sampling points is focused on the foreground area surrounding the reference point radially instead of the whole image. As for the attention weights, it is obvious that the closer the sampling point to the reference point is, the greater the attention weight will be. Specifically, the sampling points in adjacent frames are uniformly clustered in the foreground area, providing rich temporal cues of the objects in the temporal domain. This indicates that the proposed temporal deformable attention can adapt its sampling points to concentrate on the foreground object in adjacent frames.

\subsubsection{Asynchronous Inference Analysis} \label{asynchronous_analysis}
In general, the global shutter of conventional frame-based cameras limits their sampling rate, resulting in a relatively long time interval between two adjacent RGB frames. Actually, it is difficult to accurately locate moving objects with high output frequency in real-time high-speed scenarios. Towards this end, the proposed asynchronous attention-based fusion module aims at breaking through the limited output frequency from synchronized frame-based fusion methods. As illustrated in Table~\ref{tab:table11} and Fig.~\ref{fig:figure14}, we design two specific experiments to verify the effectiveness of our asynchronous fusion strategy. First, we compare the performance of single-modality baseline* without utilizing the asynchronous attention-based fusion module and our SODFormer in different frame rates. Note that we only change the frame rate, while the temporal event bins are fixed to 25 Hz, so as the output frequency. This ensures that every prediction has a corresponding annotation. We can see from Table~\ref{tab:table11} that both methods suffer a reduction in performance as the frame rate decreases, but the performance of our SODFormer is significantly less reduced than that of baseline*. Meanwhile, from the last row, we see that our SODFormer is still acceptable when the output frequency is four times the frame rate. This indicates that our asynchronous attention-based fusion module has a strong ability to conduct object detection of two asynchronous modalities. To further illustrate how our SODFormer performs in asynchronous detection when the input frequency is high, we conduct another experiment in which we use the form of a sliding window with 0.04s as its length from the continuous event stream and shift it forward by 0.01s at a time. As a result, the frequency of the divided event temporal bins can be as high as 100 Hz. Then, we input the RGB frames of 25 Hz and event temporal bins of 100 Hz into our SODFormer, as shown in Fig.~\ref{fig:figure14}. We find that the four figures in the middle show the detection performance of the event stream between the two RGB frames. Therefore, we can confirm that our asynchronous fusion strategy is able to fill the gap between two adjacent RGB frames and helps our SODFormer to detect objects in an asynchronous manner.

\begin{table}
	\caption{Comparison of the performance of SODFormer and Baseline* in different frame rates. Our Baseline* denotes SODFormer that processes only frame modality without utilizing the asynchronous attention-based fusion module.}
	\label{tab:table11}
	\vspace{-0.40cm}
	\begin{center}
    \setlength{\tabcolsep}{4.70mm}{
        \begin{tabular}{l cccc}
            \hline
				\multirow{2}*{Method} & \multicolumn{4}{c}{Frame rate (FPS)}
                \\ \cline{2-5} & 25 & 12.5 & 8.33 & 6.25 \\ 
                \toprule
                 Baseline* & 0.489 & 0.458 & 0.412 & 0.372 \\
                 \textbf{SODFormer} & 0.504 & 0.495 & 0.472 & 0.448 \\
				\hline
        \end{tabular}}
	\end{center}
	\vspace{-0.30cm}
\end{table}

\begin{figure}[h]
	\centering
	\centerline{\includegraphics[width=\linewidth]{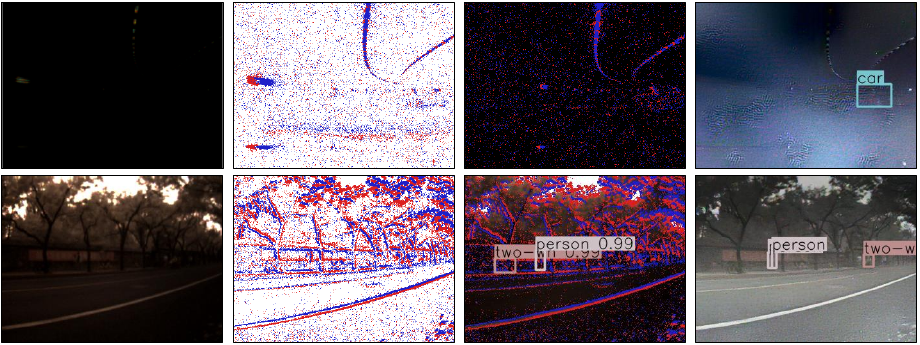}}
	\caption{Representative failure cases of our SODFormer in PKU-DAVIS-SOD dataset. The four columns from left to right refer to RGB frames, event images, our SODFormer using RGB frames and DVS events, and ground truth labeled in reconstructed images.}
	\label{fig:figure15}
	\vspace{-0.30cm}
\end{figure}

\subsubsection{Failure Case Analysis} \label{failure_case}
Although our SODFormer achieves satisfactory results even in challenging scenarios, the problem involving some failure cases is still far from being solved. As depicted in Fig.~\ref{fig:figure15}, the first row shows that static or slow-moving cars in low-light conditions are hard to perform robust detection. This is because the relatively low dynamic range of conventional cameras results in poor image quality in low-light scenes.  Meanwhile, the event camera evidently senses dynamic changes, but they generate almost no events in static or slow-moving scenarios. The second row illustrates that small moving objects at high-speed fail to be detected. This may be caused by the fact that the rushing object which is of high relative speed in RGB frames is almost invisible owing to the influence of high-speed motion blur. At the same time, event cameras capture moving objects at high-speed but they display weak textures. In fact, it indicates that our PKU-DAVIS-SOD dataset is a challenging and competitive dataset. What's more, these failure cases beyond our SODFormer need to be addressed in future work.

\section{Discussion} \label{discussion}
An effective and robust streaming object detector will further highlight the potential of the unifying framework using events and frames. Here, we will discuss the generality and the limitation of our method as follows.

\emph{Generality}. One might think that this work does not explore one core issue of how to design a novel event representation. Actually, any event representation can be regarded as the input of our SODFormer. The ablation study verifies that our SODFormer can improve detection performance by introducing DVS with different event representations (see Section~\ref{event_representation}). The highly realistic synthetic dataset should be worth exploring in the future, as they can provide large-scale data for model training and testing in simulation scenarios. However, synthetic datasets from existing simulators are unrealistic and are not suited for verifying the effectiveness of our SODFormer in extreme scenarios (e.g., low-light or high-speed motion blur). Some may argue that the backbone of our SODFormer is a CNN-based feature extractor. Indeed, this study does not design a pre-trained vision Transformer backbone (e.g., ViT~\cite{dosovitskiy2020image} and Swin Transformer~\cite{liu2021swin}) for event streams. Indeed, investigating an event-based vision Transformer to learn an effective event representation is interesting and of wide applicability in computer vision tasks (e.g., video reconstruction, object detection, and depth estimation).

\emph{Limitation}. Currently, the distribution of the sampling points in TDMSA is almost identical in different reference frames, which are radially around the reference point (see Section~\ref{visualization_attention}). However, this sampling strategy doesn't take the motion trajectory of the object into account and limits the model's capacity for modeling temporally long-term dependency (see Section~\ref{aggregation_size}). We consider following the work~\cite{zhu2017flow} of using optical flow to track the motion trajectory in optimizing the selection of sampling points, but this is not included in our SODFormer due to the additional complexity. In fact, how to track objects efficiently with a low computational complexity remains a topic worth exploring.

\section{Conclusion} \label{conclusion}
This paper presents a novel streaming object detector with Transformer (i.e., SODFormer) using events and frames, which highlights how events and frames can be utilized to deal with major object detection challenges (e.g., fast motion blur and low-light). To the best of our knowledge, this is the first trial exploring a Transformer-based architecture to continuously detect objects from two heterogeneous visual streams. To achieve this, we first build a large-scale multimodal object detection dataset (i.e., PKU-DAVIS-SOD dataset) including two visual streams and manual labels. Then, a spatiotemporal Transformer is designed to detect objects via an end-to-end sequence prediction problem, its two core innovative modules are the temporal Transformer and the asynchronous attention-based fusion module. The results demonstrate that our SODFormer outperforms four state-of-the-art methods and our eight baselines, inheriting high temporal resolution and wide HDR range properties from DVS events (i.e., brightness changes) and fine textures from RGB frames (i.e., absolute brightness). We believe that this study makes a major step towards solving the problem of fusing events and frames for streaming object detection.

\ifCLASSOPTIONcompsoc
  \section*{Acknowledgments}
\else
  \section*{Acknowledgment}
\fi

This work is partially supported by the National Natural Science Foundation of China under Grant 62027804, Grant 61825101 and Grant 62088102.

\ifCLASSOPTIONcaptionsoff
  \newpage
\fi

\bibliographystyle{IEEEtran}
\bibliography{IEEEabrv, tpami_sodformer_references}

\vspace{-1.2cm}
\begin{IEEEbiography}[{\includegraphics[width=1in,height=1.3in,clip,keepaspectratio]{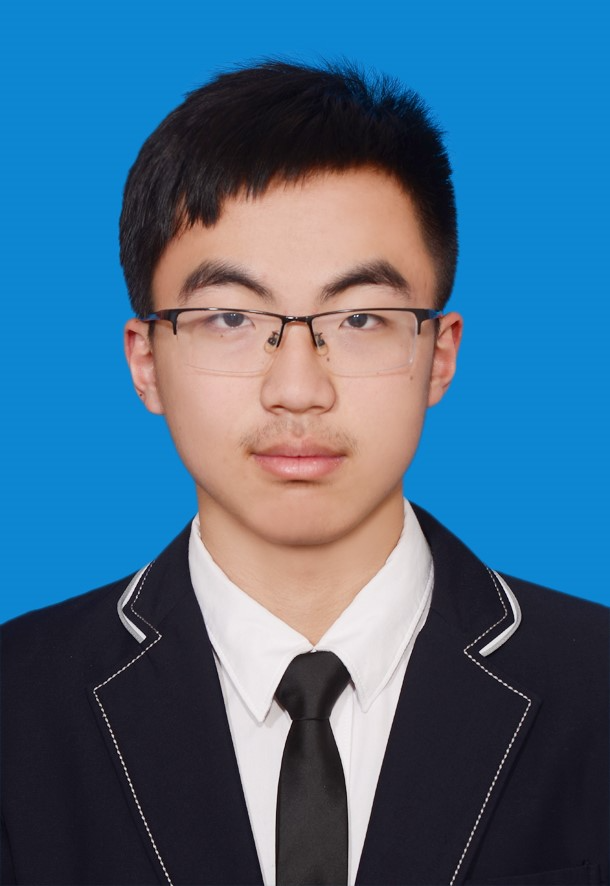}}]{Dianze Li} received the B.S degree from the School of Electronics Engineering and Computer Science, Peking University, Beijing, China, in 2022. He is currently pursuing the Ph.D. degree with the National Engineering Research Center for Visual Technology, School of Computer Science, Peking University, Beijing, China.

His current research interests include event-based vision, spiking neural networks, and neuromorphic engineering.

\end{IEEEbiography}

\vspace{-1.2cm}

\begin{IEEEbiography}[{\includegraphics[width=1in,height=1.0in,clip,keepaspectratio]{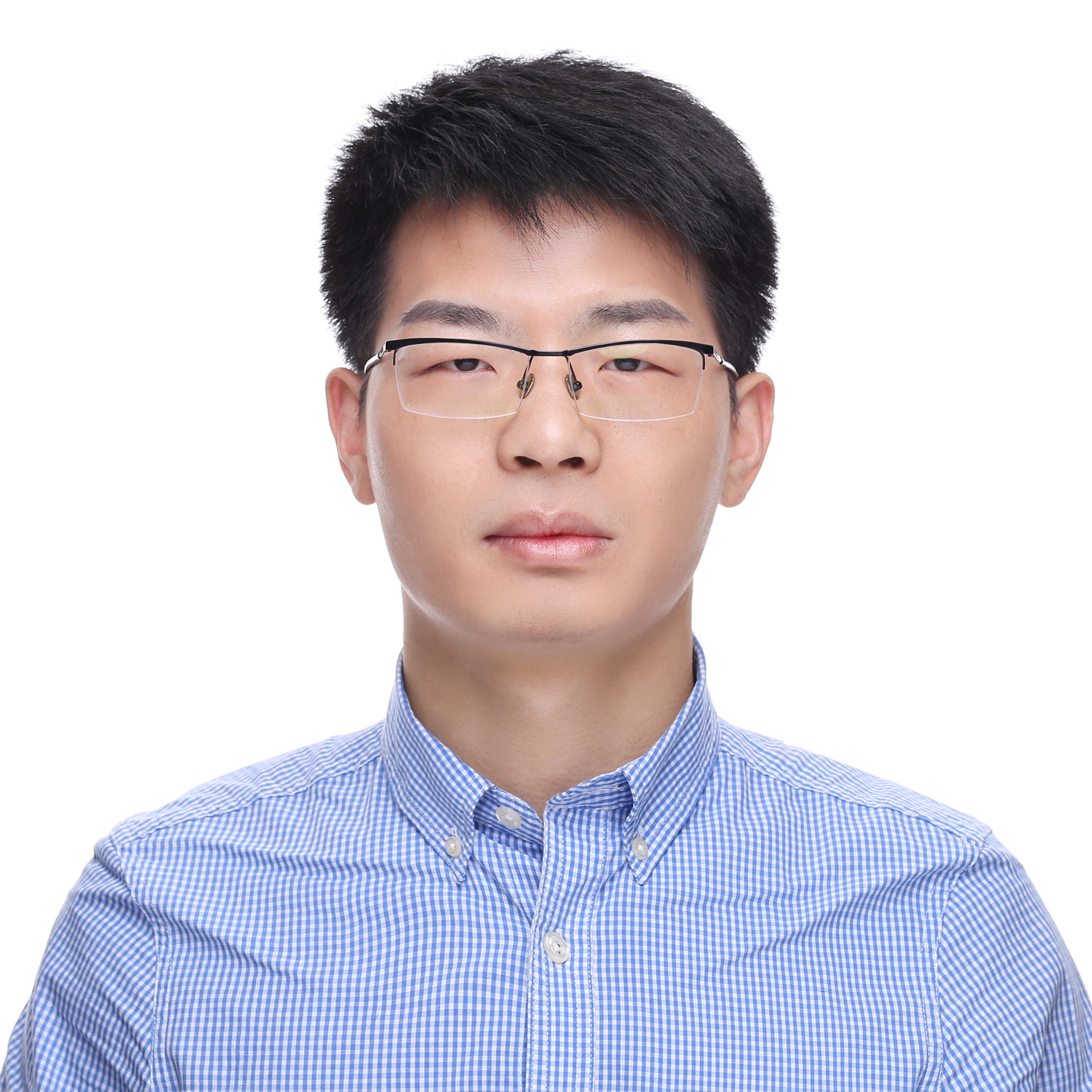}}]{Jianing Li} (Member,~IEEE) received the B.S. degree from the College of Computer and Information Technology, China Three Gorges University, China, in 2014, and the M.S. degree from the School of Microelectronics and Communication Engineering, Chongqing University, China, in 2017, and the Ph.D. degree from the National Engineering Research Center for Visual Technology, School of Computer Science, Peking University, Beijing, China, in 2022.

He is the author or coauthor of over 20 technical papers in refereed journals and conferences, such as the IEEE TRANSACTIONS ON PATTERN ANALYSIS AND MACHINE INTELLIGENCE, the IEEE TRANSACTIONS ON IMAGE PROCESSING, the IEEE TRANSACTIONS ON NEURAL NETWORKS AND LEARNING SYSTEMS, CVPR, ICCV, AAAI, and ACM MM. He received the Lixin Tang Scholarship from Chongqing University, Chongqing, China, in 2016. His research interests include event-based vision, neuromorphic engineering, and robotics.

\end{IEEEbiography}

\vspace{-1.2cm}
\begin{IEEEbiography}[{\includegraphics[width=1.2in,height=1.2in,clip,keepaspectratio]{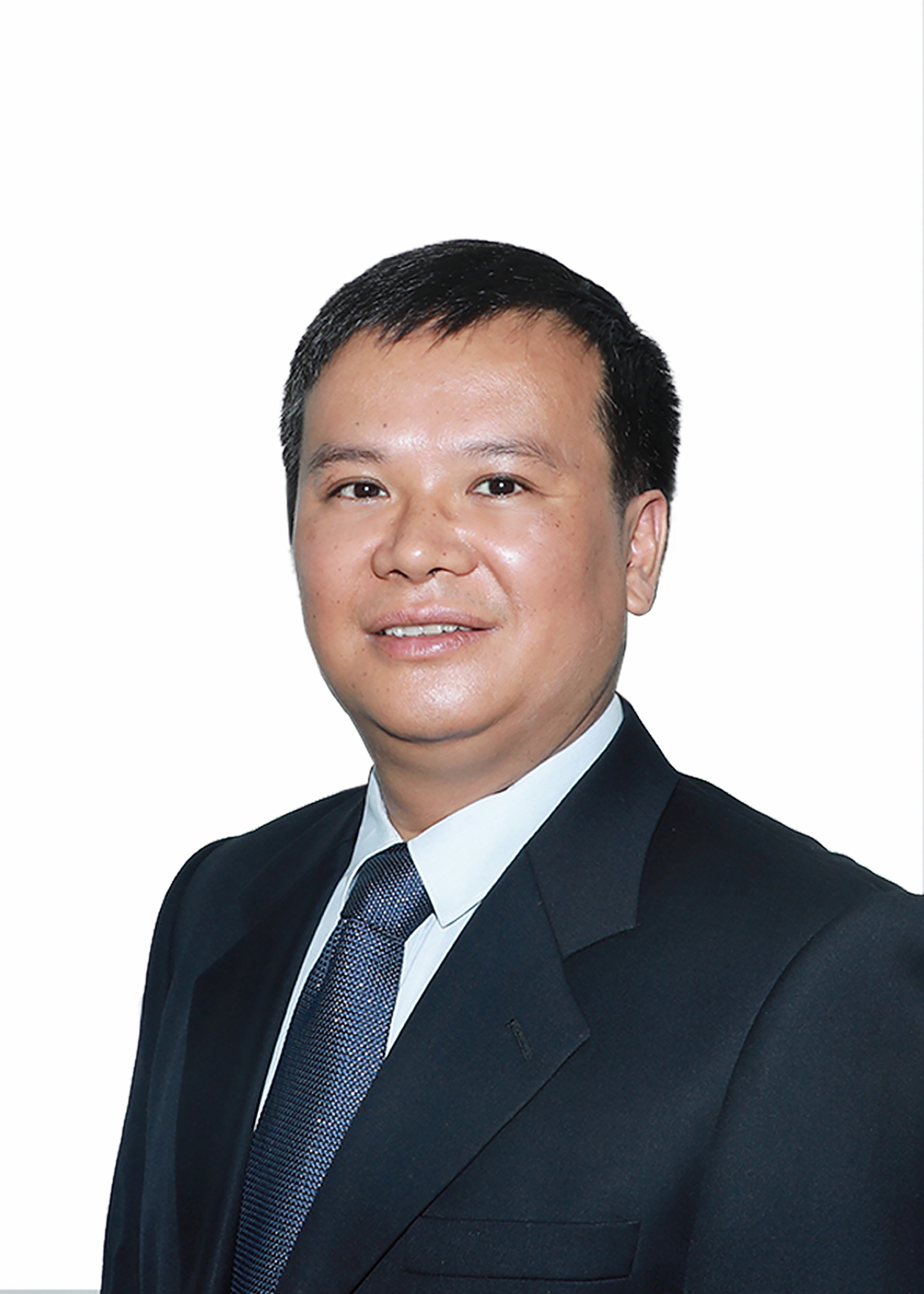}}]{Yonghong Tian} (S’00-M’06-SM’10-F’22) is currently the Dean of School of Electronics and Computer Engineering, Peking University, Shenzhen Graduate School, 518055, China, a Boya Distinguished Professor with the School of Computer Science, Peking University, China, and is also the deputy director of Artificial Intelligence Research, PengCheng Laboratory, Shenzhen, China. His research interests include neuromorphic vision, distributed machine learning and multimedia big data. He is the author or coauthor of over 300 technical articles in refereed journals and conferences. Prof. Tian was/is an Associate Editor of IEEE TCSVT (2018.1-2021.12), IEEE TMM (2014.8-2018.8), IEEE Multimedia Mag. (2018.1-2022.8), and IEEE Access (2017.1-2021.12). He co-initiated IEEE Int’l Conf. on Multimedia Big Data (BigMM) and served as the TPC Co-chair of BigMM 2015, and aslo served as the Technical Program Co-chair of IEEE ICME 2015, IEEE ISM 2015 and IEEE MIPR 2018/2019, and General Co-chair of IEEE MIPR 2020 and ICME2021. He is a TPC Member of more than ten conferences such as CVPR, ICCV, ACM KDD, AAAI, ACM MM and ECCV. He was the recipient of the Chinese National Science Foundation for Distinguished Young Scholars in 2018, two National Science and Technology Awards and three ministerial-level awards in China, and obtained the 2015 EURASIP Best Paper Award for Journal on Image and Video Processing, and the best paper award of IEEE BigMM 2018, and the 2022 IEEE SA Standards Medallion and SA Emerging Technology Award. He is a Fellow of IEEE, a senior member of CIE and CCF, a member of ACM.
\end{IEEEbiography}

\end{document}